\documentclass[10pt,twocolumn,letterpaper]{article}

\usepackage{iccv}
\usepackage{times}
\usepackage{epsfig}
\usepackage{graphicx}
\usepackage{amsmath}
\usepackage{amssymb}

\usepackage{bm}
\usepackage[T1]{fontenc}
\usepackage{multirow}
\usepackage{capt-of}

\newcommand{\bhline}[1]{\noalign{\hrule height #1}}

\renewcommand{\baselinestretch}{0.99}

\makeatletter
\newcommand{\figcaption}[1]{\def\@captype{figure}\caption{#1}}
\newcommand{\tblcaption}[1]{\def\@captype{table}\caption{#1}}
\makeatother

\usepackage[pagebackref=true,breaklinks=true,letterpaper=true,colorlinks,bookmarks=false]{hyperref}

\iccvfinalcopy

\begin{document}

\title{\vspace{-6mm}Label-Noise Robust Multi-Domain Image-to-Image Translation\vspace{-5mm}}

\author{
  Takuhiro Kaneko$^1$
  \quad Tatsuya Harada$^{1,2}$
  \vspace{2mm}\\
  $^1$The University of Tokyo
  \quad $^2$RIKEN
}

\twocolumn[{
  \renewcommand\twocolumn[1][]{#1}
  \maketitle
  \vspace{-9mm}
  \begin{center}
    \includegraphics[width=\textwidth]{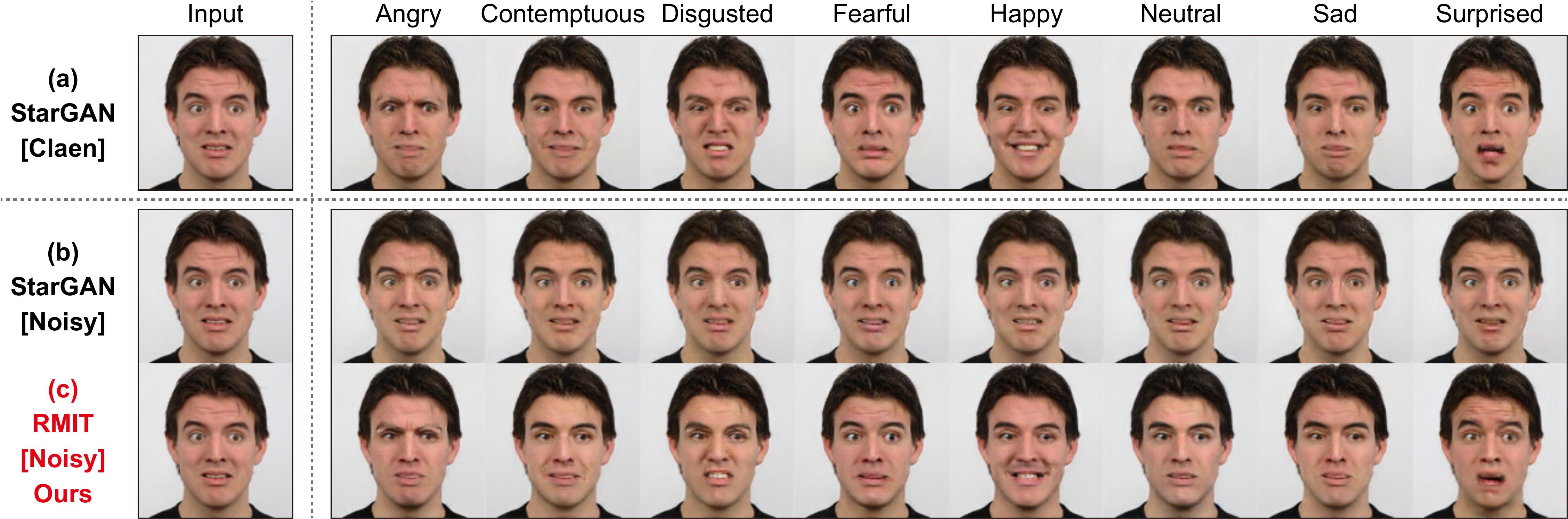}
  \end{center}
  \vspace{-3mm}
  \captionof{figure}{
    Examples of label-noise robust multi-domain image-to-image translation. The first column shows the input images while the remaining columns contain generated images. (a) Images generated by StarGAN~\cite{YChoiCVPR2018} trained using \textit{clean} labeled data. In this setting, StarGAN performs reasonably well. (b) Images generated by StarGAN trained using \textit{noisy} labeled data. In particular, we consider the situation in which training labels are flipped to the other domains with a probability of 0.5. In this setting, the noisy labels disturb StarGAN from learning meaningful conversion. (c) Images generated by our proposed RMIT trained using \textit{noisy} labeled data. Even though the training data are the same as (b), RMIT succeeds in generating images that are close to (a).}
  \label{fig:concept}
  \vspace{4mm}
}]

\begin{abstract}
  Multi-domain image-to-image translation is a problem where the goal is to learn mappings among multiple domains. This problem is challenging in terms of scalability because it requires the learning of numerous mappings, the number of which increases proportional to the number of domains. However, generative adversarial networks (GANs) have emerged recently as a powerful framework for this problem. In particular, label-conditional extensions (e.g., StarGAN) have become a promising solution owing to their ability to address this problem using only a single unified model. Nonetheless, a limitation is that they rely on the availability of large-scale clean-labeled data, which are often laborious or impractical to collect in a real-world scenario. To overcome this limitation, we propose a novel model called the label-noise robust image-to-image translation model (RMIT) that can learn a clean label conditional generator even when noisy labeled data are only available. In particular, we propose a novel loss called the virtual cycle consistency loss that is able to regularize cyclic reconstruction independently of noisy labeled data, as well as we introduce advanced techniques to boost the performance in practice. Our experimental results demonstrate that RMIT is useful for obtaining label-noise robustness in various settings including synthetic and real-world noise.
  \vspace{-5mm}
\end{abstract}

\section{Introduction}
\label{sec:introduction}
Image-to-image translation is a problem in which the goal is to translate an image into the corresponding target image. Recently, this problem has been studied actively owing to its high potential for diverse applications, such as colorization~\cite{GLarssonECCV2016,RZhangECCV2016}, super resolution~\cite{CLedigCVPR2017,WLaiCVPR2017}, image inpainting~\cite{DPathakCVPR2016,SIizukaTOG2017}, photographic image synthesis~\cite{QChenICCV2017,TCWangCVPR2018}, and photo editing~\cite{JZhuECCV2016,ABrockICLR2017,TKanekoCVPR2017}. In particular, the introduction of generative adversarial networks (GANs)~\cite{IGoodfellowNIPS2014} has resulted in significant advances in this problem and allows for an image-to-image translation model to be constructed in more challenging but practically important settings.

Among them, a well-attended problem is multi-domain image-to-image translation where the goal is to learn mapping among multiple domains. This problem focuses on a dataset that contains multiple domains, such as the RaFD dataset~\cite{OLangnerCE2010} which contains eight facial expression labels (e.g., happy, angry, and sad) and the CelebA dataset~\cite{ZLiuICCV2015} which includes 40 facial attribute labels (e.g., hair color, gender, and age). Given such a dataset, the aim of multi-domain image-to-image translation is to construct a generator that can translate an image among multiple domains according to the given domain labels (e.g., expression labels and attribute labels).

This problem is challenging in terms of scalability. In particular, typical one-to-one image-to-image translation models (e.g., \cite{YTaigmanICLR2017,TKimICML2017,JYZhuICCV2017,ZYiICCV2017,MLiuNIPS2017}) suffer from the difficulty because they require the learning of $c (c - 1)$ generators to address all mappings among the $c$ domains. To mitigate this requirement, recent studies (e.g., StarGAN~\cite{YChoiCVPR2018}) extend a conventional image-to-image translation model to the label-conditional setting. By this formulation, they enable mappings among multiple domains do be learned using only a single unified model.

Nonetheless, a possible limitation is that existing multi-domain image-to-image translation models rely on the availability of large-scale clean-labeled data, the collection of which is often laborious or impractical in a real-world scenario. Indeed, it is demonstrated that when facial expression data, which are commonly used as an application of multi-domain image-to-image translation, are collected through crowdsourcing, the annotation accuracy is low (e.g., $65 \pm 5\%$ accuracy on the FER dataset \cite{IGoodfellowNN2015}). This motivates us to address learning using noisy labeled data; however, as shown in Figure~\ref{fig:concept}(b), typical multi-domain image-to-image translation models (e.g., StarGAN in this example) are highly degraded when trained using noisy labeled data. These observations emphasize the insufficiency of the previous models. 

To overcome this limitation, we propose a \textit{label-noise robust multi-domain image-to-image translation model (RMIT)}, which can learn a \textit{clean} label conditional generator even when only \textit{noisy} labeled data are available. In particular, in StarGAN, a classification loss (which renders a generated image belong to the target domain) and cycle consistency loss (which encourages the content to be preserved during the translation) are degraded by noisy labels. To remedy this degradation, we introduce a label-noise robust classification loss and label-noise robust cycle consistency loss. Specifically, although the former has been studied actively in image classification, the latter is unique for multi-domain image-to-image translation and no established method has been devised. Hence, we propose a novel loss called the \textit{virtual cycle consistency loss} that can impose a cyclic constraint independently of noisy labeled data. Figure~\ref{fig:concept}(c) demonstrates the effectiveness of RMIT. As shown in this figure, RMIT can translate an image conditioned on \textit{clean} labels even where StarGAN is highly degraded.

Recently, a label-noise effect on DNNs has garnered attention owing to a gap between theory and practice. To reveal such a gap, empirical studies have been conducted actively in image classification~\cite{CZhangICLR2017,DArpitICML2017,DRolnickArXiv2017}; however, to our knowledge, no previous studies have analyzed such an effect on multi-domain image-to-image translation. To advance this research, we conducted extensive experiments in various label-noise settings including synthetic and real-world noise and reveal the characteristics of our novel task. Furthermore, we introduced advanced techniques for practice and empirically demonstrated their effectiveness.

Overall, our contributions are summarized as follows:
\begin{itemize}
  \vspace{-2.5mm}
  \setlength{\parskip}{1pt}
  \setlength{\itemsep}{1pt}
\item We propose a novel model called \textit{RMIT}, in which the goal is to learn a label-noise robust generator that can translate an image conditioned on \textit{clean} labels even when the training labels are \textit{noisy}.  
\item We introduce a label-noise robust classification loss and a label-noise robust cycle consistency loss into an image-to-image translation model. In particular, a label-noise robust cycle consistency loss is unique for our novel task and we devise a novel loss called the \textit{virtual cycle consistency loss}.
\item We examined the empirical performance through extensive experiments including synthetic and real-world noise along with introducing advanced techniques for practice.
  \vspace{-1.5mm}
\end{itemize}

\begin{figure*}[t]
  \centering
  \includegraphics[width=0.91\textwidth]{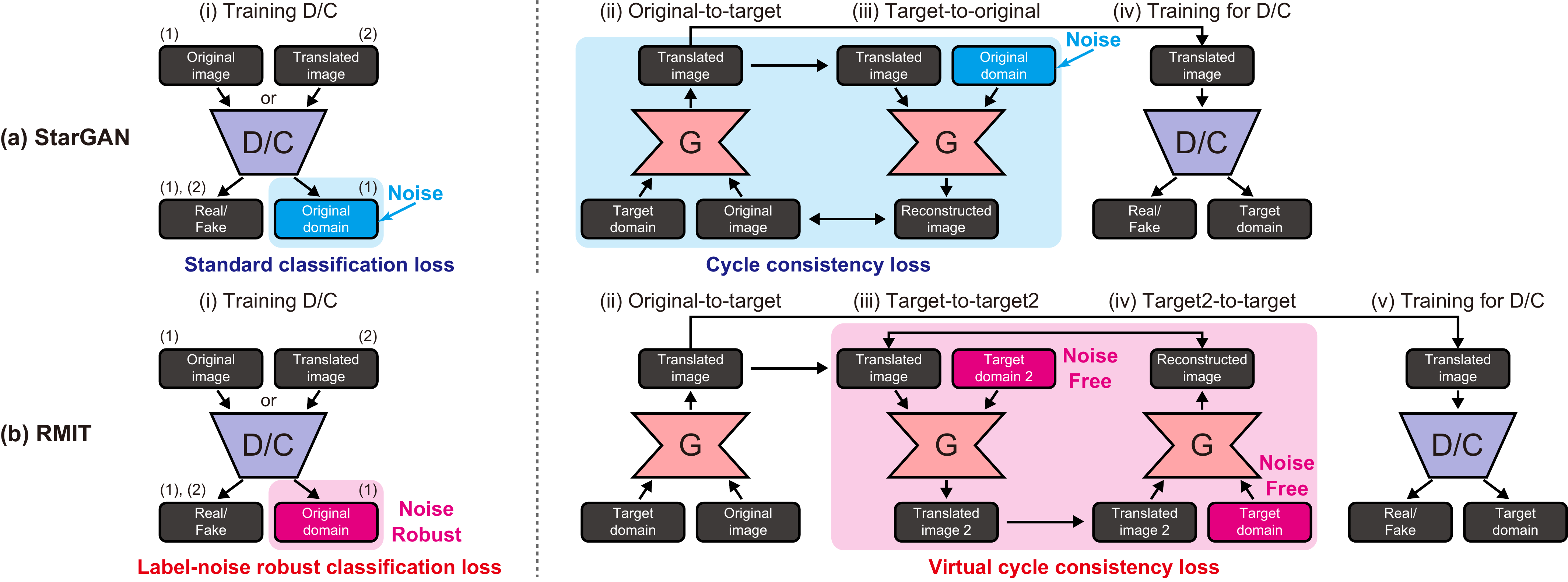}
  \caption{Overview of naive StarGAN and proposed RMIT. Both models consist of a discriminator/classifier $D/C$ and a generator $G$. The left side presents the training process of $D/C$ while the right side presents that of $G$. (a) In StarGAN, when training data are noisy, the classification loss and cycle consistency loss are problematic. (b) To alleviate this problem, we develop RMIT that incorporates a label-noise robust classification loss and a label-noise robust cycle consistency loss called the virtual cycle consistency loss.}
  \label{fig:networks}
  \vspace{-5mm}
\end{figure*}

\section{Related work}
\label{sec:related}
\noindent\textbf{Deep generative models.}
In computer vision and machine learning, generative models have been keenly studied to produce or reproduce real data. Recently, deep generative models have emerged as a powerful framework for this problem. Among them, three prominent approaches are GANs~\cite{IGoodfellowNIPS2014}, variational autoencoders (VAEs)~\cite{DKingmaICLR2014,DRezendeICML2014}, and flow-based models (Flows)~\cite{AOordICML2016,LDinhICLR2017,DPKingmaNeurIPS2018}. All these models exhibit advantages and disadvantages. Herein, we focus on GANs because they demonstrate promising results in image-to-image translation and various extensions have been proposed, as discussed in the following. One typically known problem with GANs is training instability; however, recent studies provided improvement in multiple aspects \cite{EDentonNIPS2015,ARadfordICLR2016,TSalimansNIPS2016,JZhaoICLR2017,MArjovskyICLR2017,MArjovskyICML2017,XMaoICCV2017,IGulrajaniNIPS2017,TKarrasICLR2018,XWeiICLR2018,TMiyatoICLR2018b,LMeschederICML2018,HZhangArXiv2018,ABrockICLR2019,TChenICLR2019,TKarrasArXiv2018}.

\smallskip\noindent\textbf{Conditional GANs.}
To regularize image generation, several recent studies have extended GANs to conditional settings. For example, class or attribute labels~\cite{MMirzaArXiv2014,AOdenaICML2017,TKanekoCVPR2017,ZZhangCVPR2017,TMiyatoICLR2018,TKanekoCVPR2018,YChoiCVPR2018}, texts~\cite{SReedICML2016,HZhangICCV2017,HZhangArXiv2017,XTaoCVPR2018}, object locations~\cite{SReedNIPS2016}, images~\cite{EDentonNIPS2015,PIsolaCVPR2017,CLedigCVPR2017,TCWangCVPR2018}, or videos~\cite{TCWangNeurIPS2018} are used as conditional information. This added information allows for a generator to generate a specific image that is conditioned on it. Among them, we focus on the GAN that is conditioned on both an input image and domain labels. The former is used to generate an image that is paired with an input image, and the latter is used such that a generated image follows a target domain. Such a model is typically used in the existing studies on multi-domain image-to-image translation~\cite{YChoiCVPR2018,ZHeArXiv2017,APumarolaECCV2018,BZhaoECCV2018,ARomeroArXiv2018}; however, the difference is that we address the more practical situation in which conditional labels are noisy and corrupted.

\smallskip\noindent\textbf{Image-to-image translation.}
As discussed in Section~\ref{sec:introduction}, owing to its high potential for various applications, image-to-image translation has been studied actively. In particular, GANs have broadened the applicable situations. Initially, paired image-to-image translation models~\cite{PIsolaCVPR2017,CLedigCVPR2017} have been proposed; subsequently to apply them to more practical settings, unpaired image-to-image translation models~\cite{YTaigmanICLR2017,TKimICML2017,JYZhuICCV2017,ZYiICCV2017,MLiuNIPS2017} and multi-domain image-to-image translation models~\cite{YChoiCVPR2018,ZHeArXiv2017,AAnooshehCVPRW2018,APumarolaECCV2018,BZhaoECCV2018,ARomeroArXiv2018} have been devised. To advance this research, we address \textit{label-noise robust multi-domain image-to-image translation} herein. Another popular topic is multimodal translation~\cite{JYZhuNIPS2017,AAlmahairiICML2018,XHuangECCV2018,HYLeeECCV2018}, i.e., incorporating the possibility that one input corresponds to multiple outputs. More recently, such an extension has been incorporated into multi-domain image-to-image translation models~\cite{ARomeroArXiv2018}. This model also uses a classification loss and cycle consistency loss, similar to StarGAN. Our contribution is to revise theses losses; therefore, combining our ideas into it remains a possible future direction.

\smallskip\noindent\textbf{Label-noise robust models.}
A number of studies have addressed learning with noisy labels since addressed in learning theory~\cite{DAngluinML1988,NNatarajanNIPS2013}. Recently, this problem has been addressed in image classification using DNNs. To obtain a label-noise robust classifier, a noise-tolerant loss~\cite{AGhoshAAAI2017,ZZhangNeurIPS2018}, label cleaning or sample selection methods~\cite{SReedICLR2015,DTanakaCVPR2018,EMalachNIPS2017,LJiangICML2018,MRenICML2018,BHanNeurIPS2018}, and loss correction through a noise transition model~\cite{SSukhbaatarICLRW2015,IJindalICDM2016,GPatriniCVPR2017,JGoldbergerICLR2017} have been proposed. In image generation, \textit{pixel}-noise robust models~\cite{ABoraICLR2018,JLehtinenICML2018} have begun to be studied in recent years. More recently, \textit{label}-noise robust models \cite{TKanekoCVPR2019,KKThekumparampilNeurIPS2018} have been also proposed. The primary difference is that they are \textit{image generation} models (i.e., generates an image from a random noise), while our RMIT is an \textit{image-to-image translation} model. Owing to this difference, we address a unique problem in image-to-image translation, i.e., label-noise robust cyclic reconstruction.

\section{Notations and problem statement}
\label{sec:notation}
We first define the notations and problem statement. In the following, we use superscripts $r$ and $f$ to denote the real and fake (generative) distributions, respectively. Let ${\bm x} \in {\cal X}$ be an image, and $\tilde{y} \in {\cal Y}$ and $\hat{y} \in {\cal Y}$ be the corresponding noisy (observable) and clean (unobservable) domain labels, respectively. Here, ${\cal X}$ is image space $X \subseteq \mathbb{R}^d$, where $d$ is the dimension of the image, and ${\cal Y}$ is domain label space ${\cal Y} = \{ 1, \dots, c\}$, where $c$ is the number of domains. We assume that only noisy labeled data $({\bm x}, \tilde{y}) \sim \tilde{p}^r({\bm x}, \tilde{y})$ are available during the training and clean-labeled data $({\bm x}, \hat{y}) \sim \hat{p}^r({\bm x}, \hat{y})$ cannot be accessed.

Under this condition, our aim is to learn a label-noise robust multi-domain translator $\hat{G}$ that can translate an input image ${\bm x}$ into a target-domain image ${\bm x}'$ conditioned on the \textit{clean} (but unobservable) label $\hat{y}'$, namely, $\hat{G}({\bm x}, \hat{y}') \rightarrow {\bm x}'$. This task is challenging for typical multi-domain image-to-image translation models because they are designed to fit the observable data, i.e., given noisy labeled data $({\bm x}, \tilde{y}) \sim \tilde{p}^r({\bm x}, \tilde{y})$, they attempt to learn a generator $\tilde{G}$ that translates ${\bm x}$ conditioned on the \textit{noisy} (observable) label $\tilde{y}'$, i.e., $\tilde{G}({\bm x}, \tilde{y}') \rightarrow {\bm x}'$. To solve this problem, we develop a \textit{label-noise robust multi-domain image-to-image translation model (RMIT)}. In the next section, we first briefly review StarGAN, which is the baseline of our model, and subsequently introduce our proposed RMIT.

\section{Label-noise robust multi-domain image-to-image translation: RMIT}
\label{sec:rmit}
\subsection{Background: StarGAN}
\label{subsec:stargan}
StarGAN~\cite{YChoiCVPR2018} is a prominent multi-domain image-to-image translation model and the utility of its basic idea has been shown in state-of-the-art extensions~\cite{APumarolaECCV2018,BZhaoECCV2018,ARomeroArXiv2018}. The advantage of StarGAN is that it can learn mappings among multiple domains using only a single unified model. StarGAN achieves this using three losses: an adversarial loss~\cite{IGoodfellowNIPS2014}, classification loss~\cite{AOdenaICML2017}, and cycle consistency loss~\cite{TKimICML2017,JYZhuICCV2017,ZYiICCV2017}. We present the overview of StarGAN in Figure~\ref{fig:networks}(a).

\smallskip\noindent\textbf{Adversarial loss.}
The adversarial loss~\cite{IGoodfellowNIPS2014} is used to render the generated images indistinguishable from real images:
\begin{flalign}
  \label{eqn:adv}
  {\cal L}_{adv}
  = & \: \mathbb{E}_{{\bm x} \sim p^r({\bm x})}
  [\log D({\bm x})] \nonumber \\
  + & \: \mathbb{E}_{{\bm x} \sim p^r({\bm x}), y' \sim p^f(y')}
  [\log (1 - D(G({\bm x}, y')))],
\end{flalign}
where a discriminator $D$ attempts to obtain the best decision boundary between real and generated images by maximizing this loss. In contrast, $G$ generates an image $G({\bm x}, y')$ conditioned on both the input image ${\bm x}$ and target domain label $y'$ and attempts to generate the image indistinguishable by $D$ by minimizing this loss. Here, $y'$ is sampled from $p^f(y')$ (e.g., categorical distribution ${\rm Cat} (K=c, p=\frac{1}{c})$) and independently of the real data.

\smallskip\noindent\textbf{Classification loss.}
To generate an image that belongs to the assigned domain $y'$, the classification loss~\cite{AOdenaICML2017} is introduced. First, a classifier $C$ is optimized using the classification loss of real images:
\begin{flalign}
  \label{eqn:cls_r}
  {\cal L}^r_{cls}
  = \mathbb{E}_{({\bm x}, \tilde{y}) \sim \tilde{p}^r({\bm x}, \tilde{y})}
  [- \log C(\tilde{y}|{\bm x})],
\end{flalign}
where $C(\tilde{y} | {\bm x})$ represents a probability distribution over the domain labels given ${\bm x}$. $C$ learns to classify a real image ${\bm x}$ to the corresponding domain $\tilde{y}$ by minimizing this loss.

Subsequently, $G$ is optimized using the classification loss of generated images:
\begin{flalign}
  \label{eqn:cls_f}
  {\cal L}^f_{cls}
  = \mathbb{E}_{{\bm x} \sim p^r({\bm x}), y' \sim p^f(y')}
  [- \log C(y'|G({\bm x}, y'))],
\end{flalign}
where $G$ attempts to generate an image that is classified as the target domain $y'$ by minimizing this loss.

\smallskip\noindent\textbf{Cycle consistency loss.}
The adversarial loss and classification loss only render generated images realistic and classifiable as a target domain and do not guarantee that the content is preserved between input and translated images. To alleviate this problem, the cycle consistency loss~\cite{TKimICML2017,JYZhuICCV2017,ZYiICCV2017} is used:
\begin{flalign}
  \label{eqn:cyc}
  {\cal L}_{cyc}
  = \mathbb{E}_{({\bm x}, \tilde{y}) \sim \tilde{p}^r({\bm x}, \tilde{y}), y' \sim p^f(y')}
  [\| {\bm x} - G(G({\bm x}, y'), \tilde{y}) \|_1].
\end{flalign}
This loss encourages $G$ to obtain an optimal input and target pair through cyclic reconstruction.

\smallskip\noindent\textbf{Full objective.}
In practice, the shared network between $D$ and $C$ is used. In this setting, the full objective is written as
\begin{flalign}
  \label{eqn:stargan}
  {\cal L}_{D/C} = & \: - {\cal L}_{adv} + \lambda_{cls} {\cal L}^r_{cls},
  \nonumber \\
  {\cal L}_G = & \: {\cal L}_{adv} + \lambda_{cls} {\cal L}^f_{cls}
  + \lambda_{cyc} {\cal L}_{cyc},
\end{flalign}
where $\lambda_{cls}$ and $\lambda_{cyc}$ are trade-off parameters that weigh the relative importance of the classification loss and cycle consistency loss, respectively, compared to the adversarial loss. $D$/$C$ and $G$ are optimized by minimizing ${\cal L}_{D/C}$ and ${\cal L}_G$, respectively.

\begin{figure*}[t]
  \centering
  \includegraphics[width=\textwidth]{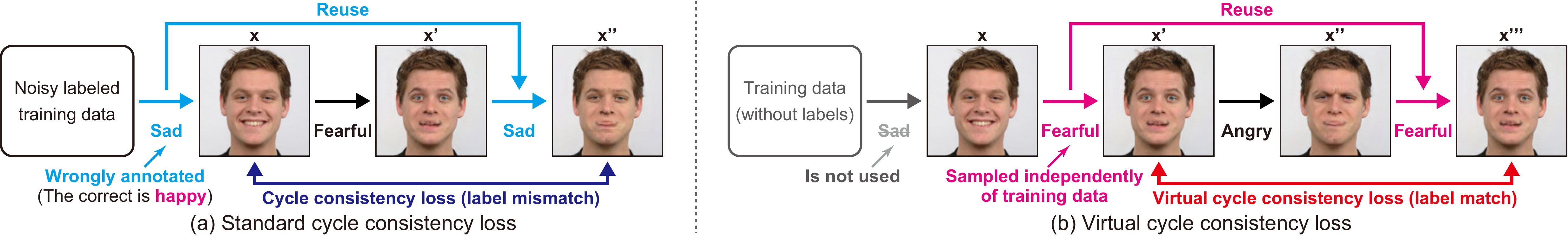}
  \caption{Comparison of standard cycle consistency loss and our proposed virtual cycle consistency loss. (a) In the standard cycle consistency loss, cyclic reconstruction is conducted from \textit{real noisy} labeled data. This causes a mismatch between the original and reconstructed images when the label is incorrect. For example, in the above, a ``happy'' image is wrongly annotated as ``sad.'' This wrong label is reused when reconstructing an image; therefore, in the cycle consistency loss, the distance between the original ``happy'' image and reconstructed ``sad'' image must be minimized. This mismatch reconstruction causes artifacts as shown in Figure~\ref{fig:recon_error}. (b) To mitigate this problem, we introduce the virtual cycle consistency loss. In this loss, reconstruction is conducted from the generated image. In this process, all labels are sampled \textit{independently of noisy labeled real} data; therefore, we can avoid the effect of the label noise. In the example above, the virtual cycle consistency loss is calculated between the images that are both labeled as ``fearful'' and label-match reconstruction is performed.}
  \label{fig:problem}
  \vspace{-5mm}
\end{figure*}

\subsection{RMIT}
\label{subsec:rmit}
In the definition above, the problematic parts when noisy labeled data are given are ${\cal L}_{cls}^r$ (Equation~\ref{eqn:cls_r}) and ${\cal L}_{cyc}$ (Equation~\ref{eqn:cyc}) because in the former, $C$ is optimized to maximize the probability distribution over noisy labeled data $C(\tilde{y} | {\bm x})$; in the latter, $G$ is optimized to conduct cyclic reconstruction conditioned on the noisy labeled data $({\bm x}, \tilde{y})$. We illustrate the position where noise is inserted, in cyan in Figure~\ref{fig:networks}(a).

To mitigate these problems, we develop RMIT that incorporates a label-noise robust classification loss and label-noise robust cycle consistency loss into StarGAN. We present the overview of RMIT in Figure~\ref{fig:networks}(b).

\smallskip\noindent\textbf{Label-noise robust classification loss.}
To obtain label-noise robustness in $C$, we replace it with a label-noise robust classifier $\hat{C}$. Although a label-noise effect on DNNs has begun to be studied in image classification~\cite{CZhangICLR2017,DArpitICML2017}, such an effect on multi-domain image-to-image translation has not been examined yet. Hence, we incorporated two different types of label-noise robust classifiers into our model and compared their performances in the experiments. The first one is \textit{forward correction}~\cite{GPatriniCVPR2017}, a widely used loss correction approach. It corrects a classification loss using a noise transition model $T = (T_{i,j}) \in [0, 1]^{c \times c}$ where $T_{i, j} = p(\tilde{y} = j | \hat{y} = i)$ represents a probability that each clean label is flipped to a noisy label. The second one is \textit{co-teaching}~\cite{BHanNeurIPS2018}, a state-of-the-art data cleaning approach. It selects out clean samples with a drop rate $\tau$ using peer classifiers. Note that a label-noise robust classifier is an orthogonal technique and any method can be incorporated into RMIT. Using either method, we reformulate ${\cal L}_{cls}^r$ and ${\cal L}_{cls}^f$ as label-noise robust ones ${\cal L}_{rcls}^r$ and ${\cal L}_{rcls}^f$, respectively.

\smallskip\noindent\textbf{Label-noise robust cycle consistency loss.}
Unlike the label-noise robust classification loss, the label-noise robust cycle consistency loss is unique for our novel task (i.e., label-noise robust multi-domain image-to-image translation), and no established method has been proposed. Therefore, we develop a novel loss called the \textit{virtual cycle consistency loss} to solve this problem. To clarify the problem and solution, we compare the standard cycle consistency loss and our proposed virtual cycle consistency loss in Figure~\ref{fig:problem}. As shown in Figure~\ref{fig:problem}(a), in the standard cycle consistency loss, cyclic reconstruction is performed from the real noisy labeled data $({\bm x}, \tilde{y}) \sim \tilde{p}^r({\bm x}, \tilde{y})$. By this definition, the cycle consistency loss suffers from a mismatch between the original and reconstructed images when the given label is incorrect. For example, in Figure~\ref{fig:problem}(a), when a ``happy'' image is wrongly labeled as ``sad,'' the difference between the original ``happy'' image and reconstructed ``sad'' image must be minimized in the cycle consistency loss. This complicates the learning of a correct image-and-label pair. Indeed, in Figure~\ref{fig:recon_error}, this mismatch reconstruction causes artifacts around the mouth.

To alleviate this problem, we develop another cycle consistency loss called the \textit{virtual cycle consistency loss} that is free from noisy labeled data. It is defined as
{
  \abovedisplayskip=9pt
  \belowdisplayskip=9pt
\begin{flalign}
  \label{eqn:vcyc}
  {\cal L}_{vcyc} = & \:
  \mathbb{E}_{{\bm x} \sim p^r({\bm x}), y' \sim p^f(y'), y'' \sim p^f(y'')}
  \nonumber \\
  & \:\:\:\:\: [\| G({\bm x}, y') - G(G(G({\bm x}, y'), y''), y') \|_1].
\end{flalign}
}
As shown in Figure~\ref{fig:problem}(b), in the virtual cycle consistency loss, the cycle consistency is considered among the \textit{virtually} generated images. Hence, we call this loss the \textit{virtual} cycle consistency loss. Unlike the cycle consistency loss (Equation~\ref{eqn:cyc}), in the virtual cycle consistency loss, labels $y'$ and $y''$ are sampled independently of the training data; hence, we can mitigate the mismatch problem caused by using noisy labeled data $({\bm x}, \tilde{y})$. For example, in Figure~\ref{fig:problem}(b), the virtual cycle consistency loss is calculated between the images that are both generated with the ``fearful'' label and label-match reconstruction is performed.

\smallskip\noindent\textbf{Full objective.}
In RMIT, the full objective is rewritten as
\begin{flalign}
  \label{eqn:rmit}
  {\cal L}_{D/C} = & \: - {\cal L}_{adv} + \lambda_{cls} {\cal L}^r_{rcls},
  \nonumber \\
  {\cal L}_G = & \: {\cal L}_{adv} + \lambda_{cls} {\cal L}^f_{rcls}
  + \lambda_{cyc} {\cal L}_{vcyc}.
\end{flalign}

\section{Advanced techniques for practice}
\label{subsec:advance}
\subsection{Relabeling technique}
\label{subsec:tech_relabel}
When a classifier is reliable enough (e.g., by using a label-noise robust classifier), another possible solution for solving the mismatch reconstruction problem is to relabel the noisy labels based on the classifier's prediction. We call this loss a \textit{relabeled cycle consistency loss} and define it as
\begin{flalign}
  \label{eqn:recyc}
  {\cal L}_{recyc} = & \:
  \mathbb{E}_{{\bm x} \sim p^r({\bm x}), y' \sim p^f(y'), y \sim C(y|{\bm x})}
  \nonumber \\
  & \:\:\:\:\:\:\:\:\:\:
  [\| {\bm x} - G(G({\bm x}, y'), y) \|_1].
\end{flalign}
Note that this loss still suffers from the mismatch problem unless a perfect clean classifier is learned, which is typically difficult in practice.

\subsection{Techniques for boosting image quality}
\label{subsec:tech_quality}
Owing to the GAN theory~\cite{IGoodfellowNIPS2014}, in an optimal condition (i.e., $D$ and $G$ exhibit sufficient capacity and the dataset is sufficiently large), it is guaranteed that a generative distribution $p^f({\bm x}')$ $({\bm x}' = G({\bm x}, y'))$ is close to a real distribution $p^r({\bm x})$, i.e., $p^f({\bm x}') = p^x({\bm x})$. Using this equation chainly, $p^f({\bm x}'')$ $({\bm x}'' = G({\bm x}', y''))$, $p^f({\bm x}''')$ $({\bm x}''' = G({\bm x}'', y''')), \dots$ also follows the real distribution as well. This confirms that even though the virtual cycle consistency loss is calculated between two generated images, it performs similar to a standard cycle consistency loss that is calculated from a real image. However, in practice, finite capacity networks are optimized using limited data. This causes the gap between real and generated images; consequently, the virtual cycle consistency loss could possibly become an inferior constraint to a cycle consistency loss. To remedy this drawback, we devised two solutions.

\smallskip\noindent\textbf{Mixed cycle consistency loss.}
The first solution is to derive the \textit{mixing loss}:
\begin{flalign}
  \label{eqn:mcyc}
  {\cal L}_{cyc\mathchar`-vcyc} = \alpha {\cal L}_{cyc} + (1 - \alpha) {\cal L}_{vcyc},
\end{flalign}
where $\alpha$ indicates the mixture rate between the cycle consistency loss and virtual cycle consistency loss. We use ${\cal L}_{cyc\mathchar`-vcyc}$ instead of ${\cal L}_{vcyc}$ in Equation~\ref{eqn:rmit}. ${\cal L}_{cyc}$ is used for regularizing the conversion based on a real image, whereas ${\cal L}_{vcyc}$ alleviates the mismatch reconstruction problem. In the experiments, we tested the variant that uses the mixing loss between ${\cal L}_{recyc}$ and ${\cal L}_{vcyc}$, i.e.,
\begin{flalign}
  \label{eqn:mrecyc}
  {\cal L}_{recyc\mathchar`-vcyc} = \alpha {\cal L}_{recyc} + (1 - \alpha) {\cal L}_{vcyc}.
\end{flalign}

\smallskip\noindent\textbf{Second adversarial loss.}
The second solution introduces the adversarial loss for the twice-converted image. We call this loss the \textit{second adversarial loss} and define it as
\begin{flalign}
  \label{eqn:adv2}
  {\cal L}_{adv2} = & \:
  \mathbb{E}_{{\bm x} \sim p^r({\bm x})}
  [ \log D'({\bm x})]
  \nonumber \\
  + & \:
  \mathbb{E}_{{\bm x} \sim p^r({\bm x}), y' \sim p^f(y'), y'' \sim p^f(y'')}
  \nonumber \\
  & \:\:\:\:\:\:\:\:\:\:
  [ \log (1 - D'(G(G({\bm x}, y'), y'')))],
\end{flalign}
where we introduce the second discriminator $D'$. $G$ is optimized by minimizing this loss while $D'$ is optimized by maximizing this loss. This loss encourages $G$ to generate a realistic image over a double conversion. We add ${\cal L}_{adv2}$ to Equation~\ref{eqn:rmit} and optimize them jointly.

\section{Experiments}
\label{sec:experiments}
To advance the research on our novel task (i.e., label-noise robust multi-domain image-to-image translation), we verified the proposed model in various settings. In Section~\ref{subsec:comprehensive}, we present a comprehensive study on RaFD~\cite{OLangnerCE2010} in diverse conditions and analyze the proposed model in detail. In Section~\ref{subsec:multiple}, we detail the testing of our model on CelebA~\cite{ZLiuICCV2015} and analyze the performance in a multi-label dataset. Finally, in Section~\ref{subsec:real}, we evaluate our model on FER~\cite{IGoodfellowNN2015} and FER+~\cite{EBarsoumICMI2016} and demonstrate the effectiveness of our model in a real-world noise setting.\footnote{Owing to space limitation, we briefly review the experimental setup and provide only the important results in this main text. See the Appendix for details and more results.}

\subsection{Comprehensive study}
\label{subsec:comprehensive}
\subsubsection{Experimental setup}
\label{subsubsec:experimental_setup}
\noindent\textbf{Dataset.}
We first performed a comprehensive study on RaFD~\cite{OLangnerCE2010} using diverse model configurations in various label-noise settings with multiple evaluation metrics. We selected this dataset because it is commonly used in multi-domain image-to-image translation (e.g., \cite{YChoiCVPR2018,APumarolaECCV2018,ARomeroArXiv2018}). This dataset consists of 4,824 images and annotated with eight facial expressions. We used $90\%$ and $10\%$ data as the training and test sets, respectively. To simulate noisy labels, we corrupted labels in two methods that are typically used in label-noise robust image classification.

\textbf{Symmetric} (class-independent) noise~\cite{RVanNIPS2015}:
For all classes, ground-truth labels are flipped to the other classes uniformly with probability $\mu \in \{ 0.25, 0.5, 0.75 \}$.

\textbf{Asymmetric} (class-dependent) noise~\cite{GPatriniCVPR2017}:
Ground-truth labels are flipped into the specific class (particularly, the next class circularly) with probability $\mu \in \{ 0.15, 0.3, 0.45 \}$.

\smallskip\noindent\textbf{Compared models.}
Our primary technical contribution is to introduce the virtual cycle consistency loss as a novel cycle consistency loss. To verify its effectiveness, we analyze the models in which the cycle consistency loss is modified.

\textbf{StarGAN:}
Naive StarGAN defined in Section~\ref{subsec:stargan}.

\textbf{StarGAN$_{recyc}$:}
StarGAN with ${\cal L}_{recyc}$ (Equation~\ref{eqn:recyc}).

\textbf{RMIT:}
Naive RMIT defined in Section~\ref{subsec:rmit}.

\textbf{RMIT$_{cyc\mathchar`-vcyc}$:}
RMIT with ${\cal L}_{cyc\mathchar`-vcyc}$ (Equation~\ref{eqn:mcyc}). In all the experiments, we set the mixture rate $\alpha = 0.5$.\footnote{We demonstrate the effect of $\alpha$ in Appendix~\ref{subsec:mixture}.}

\textbf{RMIT$_{recyc\mathchar`-vcyc}$:}
RMIT with ${\cal L}_{recyc\mathchar`-vcyc}$ (Equation~\ref{eqn:mrecyc}).

\textbf{RMIT$_{adv2}$:}
RMIT with ${\cal L}_{adv2}$ (Equation~\ref{eqn:adv2}).

\smallskip\noindent\textbf{Implementation.}
We implemented the models based on the source code provided by the authors of StarGAN.\footnote{\url{https://github.com/yunjey/StarGAN}} The network architecture is the same as that utilized in the StarGAN study~\cite{YChoiCVPR2018}: the generator network is composed of downsampling, residual~\cite{KHeCVPR2016}, and upsampling layers, as well as incorporating instance normalization~\cite{DUlyanovArXiv2016}; the discriminator network is configured as PatchGAN~\cite{CLiECCV2016}. As a GAN objective, we used CT-GAN~\cite{XWeiICLR2018}, which is a state-of-the-art GAN and an improved version of WGAN-GP~\cite{IGulrajaniNIPS2017}.

\smallskip\noindent\textbf{Training.}
Although recent studies~\cite{MLucicNeurIPS2018,KKurachArXiv2018} has demonstrated the sensitivity of GANs to hyperparameters, it is impractical or laborious to tune the hyperparameters depending on a label-noise setting when clean labels are not available. Hence, we tested the models using standard parameters, which are typically used in a clean-label setting and examined the label noise effect. Namely, we used the same setting as in the StarGAN study~\cite{YChoiCVPR2018}.

\smallskip\noindent\textbf{Evaluation metrics.}
For comprehensive analysis, we used two evaluation metrics that are typically used in multi-domain image-to-image translation or image generation.

\begin{figure*}[t]
  \def\@captype{table}
  \begin{minipage}[t]{.56\textwidth}
    \centering
    \scalebox{0.56}{
    \begin{tabular}{l|cc|cc|cc|cc|cc|cc|cc}
      \bhline{1pt}
      \multirow{3}{*}{Model}
      & \multicolumn{2}{c|}{No noise}
      & \multicolumn{6}{c|}{Symmetric noise}
      & \multicolumn{6}{c}{Asymmetric noise}
      \\ \cline{2-15}
      & \multicolumn{2}{c|}{0}
      & \multicolumn{2}{c|}{0.25}
      & \multicolumn{2}{c|}{0.5}
      & \multicolumn{2}{c|}{0.75}
      & \multicolumn{2}{c|}{0.15}
      & \multicolumn{2}{c|}{0.3}
      & \multicolumn{2}{c}{0.45}
      \\ \cline{2-15}
      & CA
      & FID
      & CA
      & FID
      & CA
      & FID
      & CA
      & FID
      & CA
      & FID
      & CA
      & FID
      & CA
      & FID
      \\ \bhline{0.75pt}
      StarGAN
      & 92.1 & \textbf{16.3}
      & 33.2 & 19.8
      & 3.2  & 20.9
      & 0.2  & 20.8
      & 72.0 & 18.3
      & 60.2 & 19.8
      & 27.7 & 18.9
      \\
      StarGAN$_{recyc}$
      & 92.3 & \textbf{16.4}
      & 36.4 & 20.2
      & 4.0  & 21.7
      & 0.3  & 21.3
      & 70.5 & 18.3
      & 57.2 & 19.0
      & 24.6 & 19.6
      \\ \hline
      RMIT
      & \textbf{92.8} & 20.1
      & \textbf{57.7} & 21.8
      & \textbf{17.3} & 20.6
      & \textbf{2.6}  & 19.8
      & \textbf{85.2} & 20.9
      & \textbf{78.9} & 21.0
      & \textbf{44.1} & 21.6
      \\
      RMIT$_{cyc\mathchar`-vcyc}$
      & 91.9 & 17.1
      & 47.1 & \textbf{19.1}
      & 6.7  & \textbf{20.5}
      & 0.5  & 19.5
      & 79.5 & \textbf{17.4}
      & 67.2 & 18.6
      & 37.3 & \textbf{18.6}
      \\
      RMIT$_{recyc\mathchar`-vcyc}$
      & 91.2 & 17.8
      & 41.8 & 19.2
      & 7.6  & 20.6
      & 0.6  & \textbf{18.0}
      & 79.7 & 17.6
      & \textbf{69.4} & \textbf{17.5}
      & \textbf{39.7} & 18.7
      \\
      RMIT$_{adv2}$
      & \textbf{94.2} & 17.4
      & \textbf{54.2} & \textbf{17.1}
      & \textbf{11.7} & \textbf{18.3}
      & \textbf{1.0}  & \textbf{18.7}
      & \textbf{84.8} & \textbf{17.1}
      & 59.1 & \textbf{17.7}
      & 30.6 & \textbf{17.0}
      \\ \bhline{1pt}
    \end{tabular}
  }
  \vspace{1mm}
  \tblcaption{Quantitative results using models without a label-noise robust classifier. The second row indicates the noise rate $\mu$. A larger CA is better, while a smaller FID is better. The two best scores are boldfaced.}
  \label{tab:score_cls}
  \vspace{-2mm}
    
  \end{minipage}
  \hfill
  \begin{minipage}[c]{.42\textwidth}

  \centering
  \includegraphics[width=0.71\textwidth]{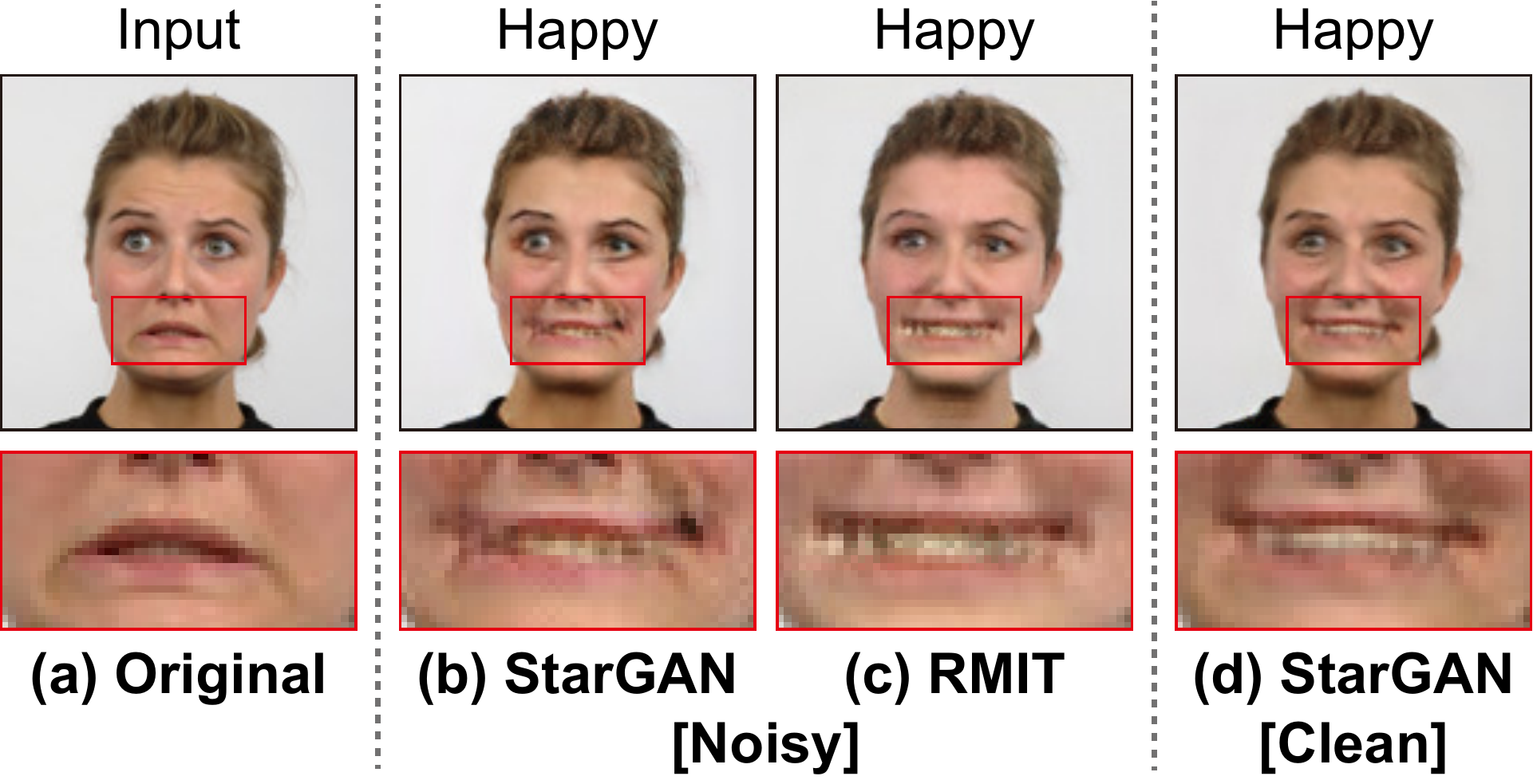}
  \vspace{-1mm}
  \caption{Generated images using models without a label-noise robust classifier (asymmetric noise with $\mu = 0.3$). StarGAN in the noisy label setting (b) contains artifacts around the mouth, while RMIT (c) generates the image that is close to that in StarGAN with the clean labels (d). See Figure~\ref{fig:recon_error_ex} for more samples.}
  \label{fig:recon_error}
\end{minipage}
\vspace{-2mm}
\end{figure*}

\begin{table*}
  \centering
  \scalebox{0.56}{
    \begin{tabular}{l|cc|cc|cc|cc|cc|cc|cc|cc|cc|cc|cc|cc}
      \bhline{1pt}
      \multirow{4}{*}{Model}
      & \multicolumn{12}{c|}{Forward correction}
      & \multicolumn{12}{c}{Co-teaching}
      \\ \cline{2-25}
      & \multicolumn{6}{c|}{Symmetric noise}
      & \multicolumn{6}{c|}{Asymmetric noise}
        & \multicolumn{6}{c|}{Symmetric noise}
      & \multicolumn{6}{c}{Asymmetric noise}
      \\ \cline{2-25}
      & \multicolumn{2}{c|}{0.25}
      & \multicolumn{2}{c|}{0.5}
      & \multicolumn{2}{c|}{0.75}
      & \multicolumn{2}{c|}{0.15}
      & \multicolumn{2}{c|}{0.3}
      & \multicolumn{2}{c|}{0.45}
      & \multicolumn{2}{c|}{0.25}
      & \multicolumn{2}{c|}{0.5}
      & \multicolumn{2}{c|}{0.75}
      & \multicolumn{2}{c|}{0.15}
      & \multicolumn{2}{c|}{0.3}
      & \multicolumn{2}{c}{0.45}
      \\ \cline{2-25}
      & CA
      & FID
      & CA
      & FID
      & CA
      & FID
      & CA
      & FID
      & CA
      & FID
      & CA
      & FID
      & CA
      & FID
      & CA
      & FID
      & CA
      & FID
      & CA
      & FID
      & CA
      & FID
      & CA
      & FID
      \\ \bhline{0.75pt}
      StarGAN
      & 75.2 & \textbf{16.5}
      & 34.3 & 17.0
      & 0.7  & \textbf{14.5}
      & 91.6 & \textbf{16.9}
      & 89.3 & \textbf{16.3}
      & 90.6 & \textbf{16.4}
      & 77.9 & 15.8
      & 44.6 & 28.6
      & 2.8  & 46.8
      & 87.7 & \textbf{15.3}
      & 81.4 & \textbf{16.0}
      & 41.4 & 26.7
      \\
      StarGAN$_{recyc}$
      & 76.1 & \textbf{16.5}
      & 32.7 & \textbf{16.7}
      & 0.7  & \textbf{14.5}
      & 91.4 & \textbf{15.9}
      & 90.6 & \textbf{16.4}
      & 90.4 & 17.0
      & 77.9 & 15.7
      & 50.2 & 27.2
      & 6.9  & 49.8
      & 89.1 & \textbf{15.3}
      & 81.8 & \textbf{15.9}
      & 49.9 & 28.0
      \\ \hline      
      RMIT
      & \textbf{86.8} & 20.3
      & \textbf{70.3} & 19.3
      & \textbf{12.9} & 19.5
      & \textbf{92.2} & 20.2
      & \textbf{92.4} & 20.2
      & \textbf{92.3} & 19.4
      & \textbf{87.6} & 18.3
      & \textbf{65.2} & \textbf{25.2}
      & \textbf{11.3} & 44.9
      & \textbf{91.7} & 17.0
      & \textbf{85.5} & 18.1
      & \textbf{66.5} & 23.1
      \\
      RMIT$_{cyc\mathchar`-vcyc}$
      & 81.0 & 17.4
      & 54.4 & 17.0
      & 6.1  & 15.4
      & 90.9 & 17.0
      & 90.7 & 17.3
      & 90.6 & \textbf{16.9}
      & 81.3 & \textbf{15.6}
      & 53.5 & 27.7
      & 11.0 & \textbf{39.8}
      & 90.2 & 15.8
      & 85.4 & 16.5
      & 45.7 & 24.8
      \\
      RMIT$_{recyc\mathchar`-vcyc}$
      & 81.3 & 16.8
      & 62.6 & \textbf{16.5}
      & 7.8  & 15.6
      & 91.2 & \textbf{16.9}
      & 90.4 & 17.2
      & 90.5 & \textbf{16.9}
      & 82.1 & \textbf{15.6}
      & 56.1 & \textbf{25.3}
      & \textbf{12.1} & \textbf{44.0}
      & 90.3 & \textbf{15.3}
      & 84.5 & 16.4
      & \textbf{53.2} & \textbf{22.4}
      \\
      RMIT$_{adv2}$
      & \textbf{88.0} & 17.3
      & \textbf{69.5} & 18.1
      & \textbf{12.5} & 17.3
      & \textbf{92.9} & 18.8
      & \textbf{93.8} & 16.9
      & \textbf{93.6} & 17.6
      & \textbf{89.3} & 15.7
      & \textbf{62.9} & 27.5
      & 7.0  & 55.8
      & \textbf{92.6} & 16.0
      & \textbf{89.7} & 16.1
      & 50.0 & \textbf{21.9}
      \\ \bhline{1pt}
    \end{tabular}
  }
  \vspace{1mm}
  \caption{Quantitative results using models with label-noise robust classifiers. The third row indicates the noise rate $\mu$. A larger CA is better, while a smaller FID is better. The two best scores are boldfaced.}
  \label{tab:score_rcls}
  \vspace{-5mm}
\end{table*}

\textbf{Classification accuracy (CA)}~\cite{YChoiCVPR2018,BZhaoECCV2018}:
To evaluate whether a translated image belongs to the correct target domain, we used the CA. We first trained a classifier (in particular, we used PreAct ResNet-18~\cite{KHeECCV2016}) using clean-labeled training data. We trained it independently of image translation models. We subsequently translated images in the test set to the domain that is different from the original domain using each image translation model. Finally, we calculated the accuracy of the translated images using the above-mentioned classifier. By this definition, when a nonconversion model is learned (such a model tends to be learned in a severely noisy case), the CA becomes close to $0 \%$.

\textbf{Fr\'{e}chet Inception distance (FID)}~\cite{MHeuselNIPS2017}:
To evaluate the fidelity of the generated images, we used the FID, which measures the distance between real and generated data in the Inception embeddings~\cite{CSzegedyCVPR2016}. Precisely, we first translated an image in the test set using the labels of another image. We subsequently calculated the FID between the translated samples and all the real samples in the training set.

It is noteworthy that only achieving a low FID is problematic in our task because a nonconversion model or completely noisy-label-fitting model, which tend to be learned in highly noisy cases, can also achieve a low FID. It is important to obtain a good score in both CA and FID.
Other popular metrics are the Inception score (IS)~\cite{TSalimansNIPS2016} and Kernel Inception distance (KID)~\cite{MBinkowskiICLR2018}. Recent studies show that IS has several drawbacks~\cite{MHeuselNIPS2017,MLucicNeurIPS2018,SBarrattICMLW2018} and KID is correlated with FID in terms of ranking~\cite{KKurachArXiv2018}. As reference, we discuss the correlation among evaluation metrics in Appendix~\ref{subsec:analysis_eval}.

\subsubsection{Comparison using naive classifier}
\label{subsubsec:naive_cls}
To examine the effectiveness of virtual consistency loss itself, we first analyzed the performance without a label-noise robust classifier. We list the quantitative results in Table~\ref{tab:score_cls}. In the pre-experiments, we found that an extremely bad or good initialization causes an outlier, possibly because of the instability of GAN training or the ambiguity caused by label noise. Hence, we report the median value over five trials.

Regarding the CA, RMIT achieved the best performance in most cases. This verifies that the virtual cycle consistency loss is useful for resisting the label noise. However, regarding the FID, RMIT exhibits poor scores in some cases. As discussed in Section~\ref{subsec:advance}, this is possibly because the virtual cycle consistency loss is calculated between the generated images and is weak compared to the cycle consistency loss stemming from real images. However, this degradation is not reflected in the advanced RMIT (RMIT$_{cyc\mathchar`-vcyc}$, RMIT$_{recyc\mathchar`-vcyc}$, and RMIT$_{adv2}$) while maintaining a better CA than StarGAN. Between StarGAN and StarGAN$_{recyc}$, a better or worse performance is case dependent (the difference in the CA is almost within three). This is because a naive classifier can easily memorize noisy labels without any regularization~\cite{CZhangICLR2017,DArpitICML2017}, and the labels relabeled by the classifier are close to the original noisy labels.

We show generated images in Figure~\ref{fig:recon_error}. StarGAN with the noisy labels (b) contains artifacts around the mouth. As discussed in Figure~\ref{fig:problem}, StarGAN suffers from the mismatch reconstruction problem. This disturbs the generator from learning a completely domain-specific image. In contrast, this problem is resolved in RMIT (c) and the generated image is close to that in StarGAN with the clean labels (d).

\subsubsection{Comparison using label-noise robust classifiers}
\label{subsubsec:robust_cls}
We subsequently examined the performance when using the label-noise robust classifiers jointly. In particular, we tested two types of label-noise classifiers: \textit{forward correction}~\cite{GPatriniCVPR2017} and \textit{co-teaching}~\cite{BHanNeurIPS2018}, as described in Section~\ref{subsec:rmit}. Regarding forward correction, we assume that the noise transition matrix $T$ is known; regarding co-teaching, we set the drop rate $\tau$ the same value as the noise rate $\mu$.

We list the quantitative results in Table~\ref{tab:score_rcls}. We observed a similar tendency in Table~\ref{tab:score_cls}. RMIT achieves the best CA in most cases while it tends to degrade the FID. However, it is recovered by the advanced RMIT while maintaining a higher CA. This confirms that the virtual cycle consistency loss is useful for our task even when label-noise robust classifiers are available.\footnote{To further confirm these claims, we summarize the results across all the conditions in Appendix~\ref{subsec:summarization}.} When comparing StarGAN and StarGAN$_{recyc}$, we found that in some cases, the CA is improved by a large margin (e.g., 8.5 in co-teaching with asymmetric noise ($\mu = 0.45$)). This indicates that when a classifier is reliable enough, a simple relabeling is also useful. This tendency is also observed in the comparison between RMIT$_{cyc\mathchar`-vcyc}$ and RMIT$_{recyc\mathchar`-vcyc}$. The comparison between forward correction and co-teaching indicates that forward correction tends to outperform co-teaching. Note that forward correction requires a strong assumption (i.e., a noise transition matrix $T$ is known), while co-teaching relies on a relatively weak assumption (i.e., only the drop rate $\tau$ needs to be set). In co-teaching, the FID is highly degraded when the noise rate $\mu$ is high. One possible reason is that sample selection by co-teaching disturbs the generator from covering all data distributions.

\subsection{Application to multi-label dataset}
\label{subsec:multiple}
A multi-label dataset is a primary target of a multi-domain image translation model. To confirm the effectiveness in such a dataset, we conducted experiments on CelebA~\cite{ZLiuICCV2015}. Similar to the StarGAN study~\cite{ZLiuICCV2015}, we choose five attributes: three hair colors (black, blond, and brown), gender, and age. We flipped each label independently with the noise rate $\mu \in \{ 0.3, 0.45 \}$. In this case, using a label-noise robust classifier is impractical because the number of parameters (e.g., $T$ in forward correction and $\tau$ in co-teaching) increases depending on the number of attributes. Hence, we tested the models using a naive classifier. In Section~\ref{subsec:comprehensive}, we found that the relabeled cycle consistency loss is not effective without a label-noise robust classifier; therefore, we excluded it for comparison.

\begin{figure}[t]
  \centering
  \includegraphics[width=\columnwidth]{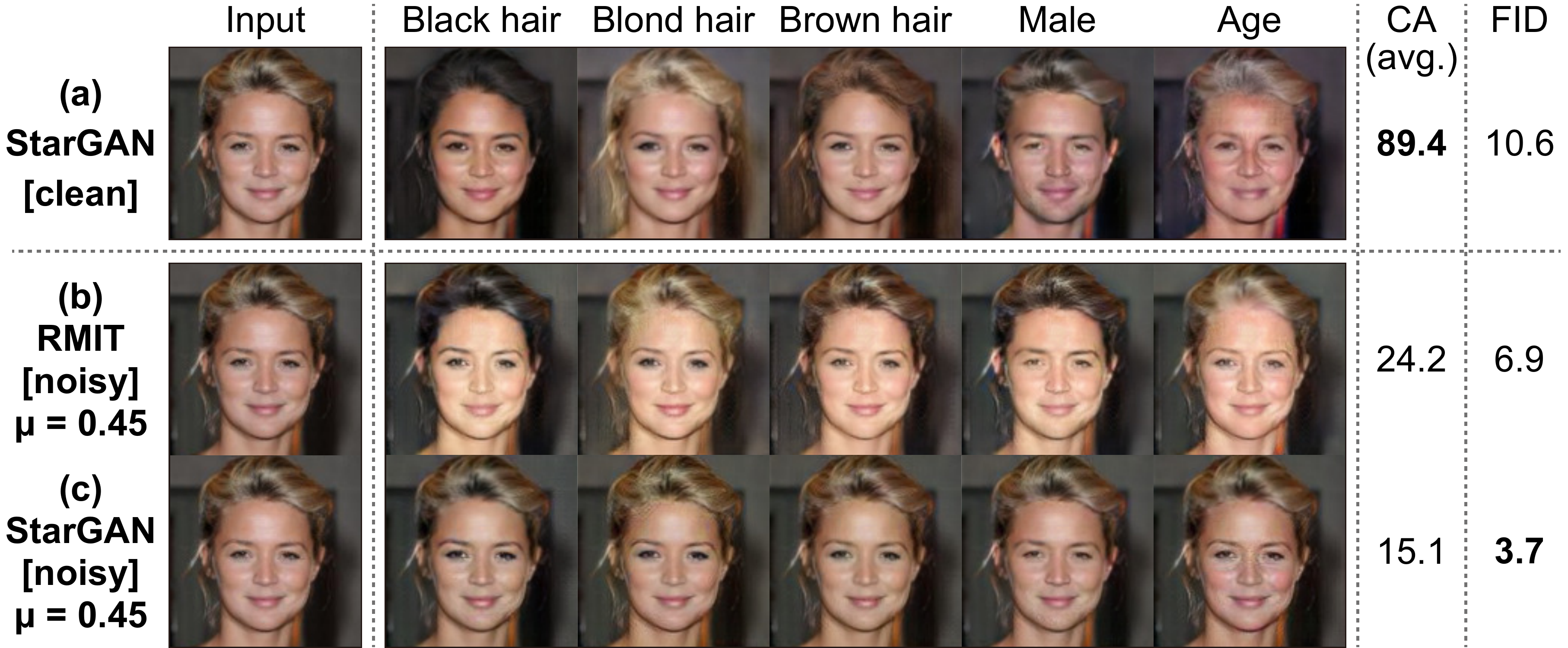}
  \caption{Generated images on CelebA. In the noisy label setting, (b) RMIT struggles to conduct meaningful conversion while (c) StarGAN is close to a nonconversion model. It is noteworthy that in the upper rows, the CA is better but the FID is worse because a nonconversion model is preferable in terms of the FID in this dataset. See Figure~\ref{fig:celeba_ex} for more samples.}
  \label{fig:celeba}
  \vspace{-6mm}
\end{figure}

We list the quantitative results in Table~\ref{tab:score_celeba}. We report the median value over three trials. We calculated the CA for the images in which a single attribute of either hair color, gender, or age is translated. Similar to the experiments in Section~\ref{subsec:comprehensive}, we translated an image to the different domain from the original; therefore, when a nonconversion model is learned, the CA becomes close to $0 \%$. These results confirm that RMIT achieved the best CA and RMIT$_{cyc\mathchar`-vcyc}$ obtained the balancing scores. We found that RMIT$_{adv2}$ did not perform well in this case possibly because balancing between two discriminators becomes difficult depending on a dataset. Improving this remains a possible future work. Another finding is that the FID tends to become smaller as the noise rate increases. This is because the FID score is preferred by a nonconversion model (which can achieve the FID of 1.1). Generated images shown in Figure~\ref{fig:celeba} confirm this claim. This finding indicates that balancing between the FID and CA is important for achieving good label-noise robust multi-domain image-to-image translation.

\begin{table}
  \centering
  \scalebox{0.56}{
    \begin{tabular}{l|ccc|c|ccc|c|ccc|c}
      \bhline{1pt}
      \multirow{3}{*}{Model}
      & \multicolumn{4}{c|}{Clean}
      & \multicolumn{4}{c|}{Noisy ($\mu = 0.3$)}
      & \multicolumn{4}{c}{Noisy ($\mu = 0.45$)}
      \\ \cline{2-13}
      & \multicolumn{3}{c|}{CA}
      & FID
      & \multicolumn{3}{c|}{CA}
      & FID
      & \multicolumn{3}{c|}{CA}
      & FID
      \\
      & H & G & A &
      & H & G & A &
      & H & G & A &
      \\ \bhline{0.75pt}
      StarGAN
      & 88.0
      & 93.3
      & 87.5
      & \textbf{10.6}
      & 63.6
      & 71.3
      & 58.5
      & \textbf{9.8}
      & 9.0
      & 8.8
      & 27.9
      & \textbf{3.7}
      \\ \hline
      RMIT
      & \textbf{89.5}
      & \textbf{97.0}
      & \textbf{90.0}
      & 14.0
      & \textbf{68.5}
      & \textbf{83.2}
      & \textbf{65.8}
      & 14.8
      & \textbf{19.4}
      & \textbf{19.2}
      & \textbf{31.5}
      & 6.9
      \\
      RMIT$_{cyc\mathchar`-vcyc}$
      & \textbf{88.2}
      & \textbf{95.9}
      & \textbf{88.1}
      & \textbf{10.0}
      & \textbf{64.1}
      & \textbf{76.2}
      & \textbf{59.9}
      & \textbf{10.3}
      & \textbf{12.9}
      & \textbf{11.1}
      & \textbf{28.7}
      & \textbf{4.0}
      \\
      RMIT$_{adv2}$
      & 84.2
      & 93.6
      & 87.0
      & 17.4
      & 55.8
      & 67.1
      & 59.9
      & 12.6
      & 6.0
      & 3.3
      & 22.1
      & 11.2
      \\ \bhline{1pt}
    \end{tabular}
  }
    \vspace{0.5mm}
  \caption{Quantitative results on CelebA. The first row indicates the noise rate $\mu$. 'H,' 'G,' and 'A' denote hair color, gender, and age, respectively. A larger CA is better, while a smaller FID is better. The two best scores boldfaced.}
  \label{tab:score_celeba}
  \vspace{-2mm}
\end{table}

\subsection{Evaluation on real-world noisy dataset}
\label{subsec:real}

Finally, we evaluated the models on FER~\cite{IGoodfellowNN2015} and FER+~\cite{EBarsoumICMI2016} to verify the effectiveness on a real-world noisy dataset. These two datasets contain the same images; however, their annotation methods are different. The original FER is created by web crawling with emotion-related keywords. Although the images are filtered by human labels, it is shown that the label accuracy is not high ($65 \pm 5 \%$)~\cite{IGoodfellowNN2015}. To increase the label accuracy, in FER+, each image is relabeled by 10 crowd-sourced taggers. It is shown that by increasing the number of taggers, the agreement percentage can be improved~\cite{EBarsoumICMI2016}. We regard the labels in FER as noisy labels and the majority-voting labels in FER+ as clean labels. We chose five emotions (neutral, happy, surprised, sad, and angry) because the number of samples in the other classes is low. In the pre-experiments, we observed that a standard network and training setting does not perform well possibly because this dataset is gray and not well aligned. However, we found that an identity mapping loss~\cite{YTaigmanICLR2017,JYZhuICCV2017} and attention mechanisms~\cite{APumarolaECCV2018} are useful for solving this problem. Therefore, we applied them in the experiments. We examined the performance using a standard classifier.

\begin{figure}[t]
  \centering
  \includegraphics[width=0.7\columnwidth]{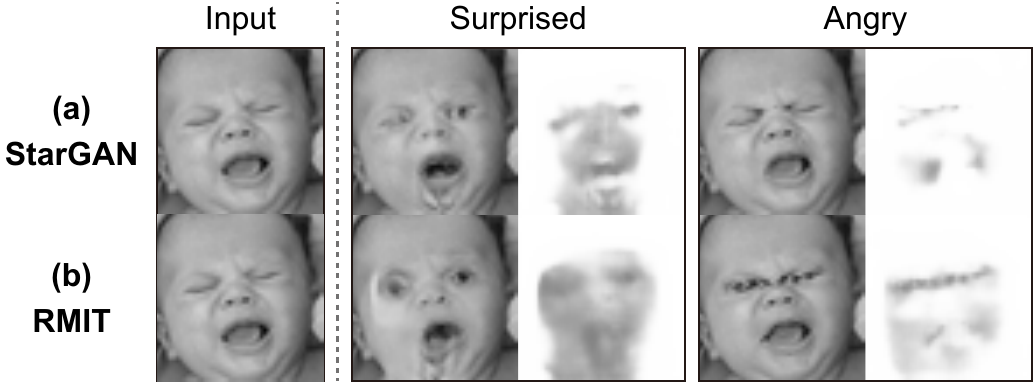}
  \caption{Generated images on FER. The third and fifth columns show attention masks. RMIT (b) tends to generate more classifiable images than StarGAN (a). See Figure~\ref{fig:fer_ex} for more samples.}
  \label{fig:fer}
  \vspace{-6mm}
\end{figure}

We list the quantitative results in Table~\ref{tab:score_fer} in which we report the median value over five trials. Although there is a gap between the models learned in the clean settings and those in the noisy settings, the results confirmed that even in the real-world noise setting, RMIT is useful for improving the CA, and that RMIT$_{cyc\mathchar`-vcyc}$ and RMIT$_{adv2}$ achieved the balancing performance. We show the generated images and attention masks in Figure~\ref{fig:fer}.

\begin{table}
  \centering
  \scalebox{0.56}{
    \begin{tabular}{l|cc|cc}
      \bhline{1pt}
      \multirow{2}{*}{Model}
      & \multicolumn{2}{c|}{FER+ (clean)}
      & \multicolumn{2}{c}{FER (noisy)}
      \\ \cline{2-5}
      & CA & FID
      & CA & FID
      \\ \bhline{0.75pt}
      StarGAN
      & 76.3 & \textbf{6.6}
      & 65.5 & \textbf{6.8}
      \\ \hline
      RMIT
      & \textbf{81.5} & 7.2
      & \textbf{70.0} & 7.5
      \\
      RMIT$_{cyc\mathchar`-vcyc}$
      & \textbf{80.3} & 6.9
      & 67.7 & 7.1
      \\
      RMIT$_{adv2}$
      & 79.8 & \textbf{6.7}
      & \textbf{67.8} & \textbf{6.8}
      \\ \bhline{1pt}
    \end{tabular}
  }
  \vspace{1mm}
  \caption{Quantitative results on FER and FER+. A larger CA is better, while a smaller FID is better. Best two scores are boldfaced.}
  \label{tab:score_fer}
  \vspace{-4mm}
\end{table}

\vspace{-0.5mm}
\section{Conclusion}
\label{sec:conclusion}
Recently, variants of multi-domain image-to-image translation models have demonstrated promising results; however, they require the access to the large-scale clean-labeled data. To overcome this limitation, we developed a novel model called RMIT. In particular, we devised a novel loss called the virtual cycle consistency loss along with several advanced techniques for practice. Our experimental results demonstrated the effectiveness of RMIT in various settings. Our proposed techniques were orthogonal to various extensions of multi-domain image-to-image translation models (e.g., stochastic extension~\cite{ARomeroArXiv2018} or introduction of continuous supervision~\cite{APumarolaECCV2018}). Incorporating our idea into them remains an interesting future direction.

\renewcommand{\baselinestretch}{1}\selectfont
\section*{Acknowledgement}
This work was partially supported by JST CREST Grant Number JPMJCR1403, Japan, and partially supported by the Ministry of Education, Culture, Sports, Science and Technology (MEXT) as ``Seminal Issue on Post-K Computer.''

{\small
\bibliographystyle{ieee}
\bibliography{refs}
}

\clearpage
\onecolumn
\appendix
\section{Additional analysis}
\label{sec:add_results}

\subsection{Summarization of results}
\label{subsec:summarization}

Our main two claims in the comprehensive study in Section~\ref{subsec:comprehensive} are as follows:
\begin{itemize}
  \vspace{-1mm}
  \setlength{\parskip}{0pt}
  \setlength{\itemsep}{0pt}
\item Regarding the classification accuracy (CA), RMIT outperforms StarGAN in most cases. However, in terms of the Fr\'{e}chet Inception distance (FID), RMIT exhibits poor scores in some cases.
\item The above degradation is not reflected in the advanced RMITs (i.e., RMIT$_{cyc\mathchar`-vcyc}$, RMIT$_{recyc\mathchar`-vcyc}$, and RMIT$_{adv2}$). These models can achieve a comparable performance to StarGAN in terms of the FID while maintaining a better CA than StarGAN.
  \vspace{-1mm}
\end{itemize}
In this section, we summarize the results across all the conditions, and provide their statistics to confirm these claims.

\subsubsection{Experimental conditions}
In Table~\ref{tab:model}, we summarize the differences in the generator objectives among the six compared models in Section~\ref{subsec:comprehensive}. Except for RMIT$_{adv2}$, only the cycle consistency loss term is different, and the other terms are the same. In contrast, in RMIT$_{adv2}$, the second adversarial loss ${\cal L}_{adv2}$ is additionally utilized. Similarly, except for RMIT$_{adv2}$, the same discriminator objective is utilized. In RMIT$_{adv2}$ only, the second discriminator $D'$ is simultaneously optimized.

In Table~\ref{tab:condition}, we list the 19 conditions analyzed in Section~\ref{subsec:comprehensive}. These include both the seven conditions used in Section~\ref{subsubsec:naive_cls} (i.e., comparison using a naive classifier) and the 12 used in Section~\ref{subsubsec:robust_cls} (i.e., comparison using label-noise robust classifiers). To mitigate the effect of initialization, we trained each model with each condition five times with different random initializations. Hence, we trained $6 \:({\rm models}) \times 19 \:({\rm conditions}) \times 5 \:({\rm initializations})= 570$ models in total in these experiments.

\begin{table}[h]
  \vspace{-1mm}
  \centering  
  \begin{tabular}{cc}
    \begin{minipage}[t]{.48\linewidth}
      \scalebox{0.8}{
        \begin{tabular}{l|l}          
          \bhline{1pt}
          \multirow{2}{*}{Model}
          & \multirow{2}{*}{Generator objective}
          \\
          \\ \bhline{0.75pt}
          StarGAN
                & ${\cal L}_{adv}
                  + \lambda_{cls} {\cal L}_{cls}^f
                  + \lambda_{cyc} {\cal L}_{cyc}$
          \\
          StarGAN$_{recyc}$
                & ${\cal L}_{adv}
                  + \lambda_{cls} {\cal L}_{cls}^f
                  + \lambda_{cyc} {\cal L}_{recyc}$
          \\ \hline
          RMIT
                & ${\cal L}_{adv}
                  + \lambda_{cls} {\cal L}_{cls}^f
                  + \lambda_{cyc} {\cal L}_{vcyc}$
          \\
          RMIT$_{cyc\mathchar`-vcyc}$
                & ${\cal L}_{adv}
                  + \lambda_{cls} {\cal L}_{cls}^f
                  + \lambda_{cyc} ( \alpha {\cal L}_{cyc}
                  + (1 - \alpha) {\cal L}_{vcyc})$
          \\
          RMIT$_{recyc\mathchar`-vcyc}$
                & ${\cal L}_{adv}
                  + \lambda_{cls} {\cal L}_{cls}^f
                  + \lambda_{cyc} ( \alpha {\cal L}_{recyc}
                  + (1 - \alpha) {\cal L}_{vcyc})$
          \\
          RMIT$_{adv2}$
                & ${\cal L}_{adv}
                  + \lambda_{cls} {\cal L}_{cls}^f
                  + \lambda_{cyc} {\cal L}_{vcyc}
                  + {\cal L}_{adv2}$
          \\ \bhline{1pt}
        \end{tabular}
      }
      \vspace{1mm}
      \caption{Six models compared in the comprehensive study in Section~\ref{subsec:comprehensive}. The right column includes the corresponding generator objectives. Except for RMIT$_{adv2}$, only the cycle consistency loss term is different, and the other terms are the same. In RMIT$_{adv2}$ only, the second adversarial loss ${\cal L}_{adv2}$ is additionally utilized.}
      \label{tab:model}
    \end{minipage}

    \hfill{\hspace{3mm}}

    \begin{minipage}[t]{.48\linewidth}
      \scalebox{0.8}{
        \begin{tabular}{c|cc|c}
          \bhline{1pt}
          \multirow{2}{*}{Classifier}
          & \multirow{2}{*}{Noise type}
          & \multirow{2}{*}{Noise rate} & No. of
          \\
          & & & conditions
          \\ \bhline{0.75pt}
          \multirow{3}{*}{Naive} & No & --
          \\
          & Symmetric & $\mu \in \{ 0.25, 0.5, 0.75 \}$ & 7
          \\
          & Asymmetric & $\mu \in \{ 0.15, 0.3, 0.45 \}$
          \\ \hline
          Forward & Symmetric & $\mu \in \{ 0.25, 0.5, 0.75 \}$ & \multirow{2}{*}{6}
          \\
          correction & Asymmetric & $\mu \in \{ 0.15, 0.3, 0.45 \}$
          \\ \hline
          \multirow{2}{*}{Co-teaching} & Symmetric & $\mu \in \{ 0.25, 0.5, 0.75 \}$
                                        & \multirow{2}{*}{6}
          \\
          & Asymmetric & $\mu \in \{ 0.15, 0.3, 0.45 \}$
          \\ \bhline{1pt}
        \end{tabular}
      }
      \vspace{1mm}
      \caption{Nineteen conditions analyzed in the comprehensive study in Section~\ref{subsec:comprehensive}.}
      \label{tab:condition}
    \end{minipage}
  \end{tabular}
  \vspace{-5mm}
\end{table}

\subsubsection{Comparison results}
\begin{figure*}[h]
  \vspace{-2mm}
  \centering
  \includegraphics[height=0.773\textheight]{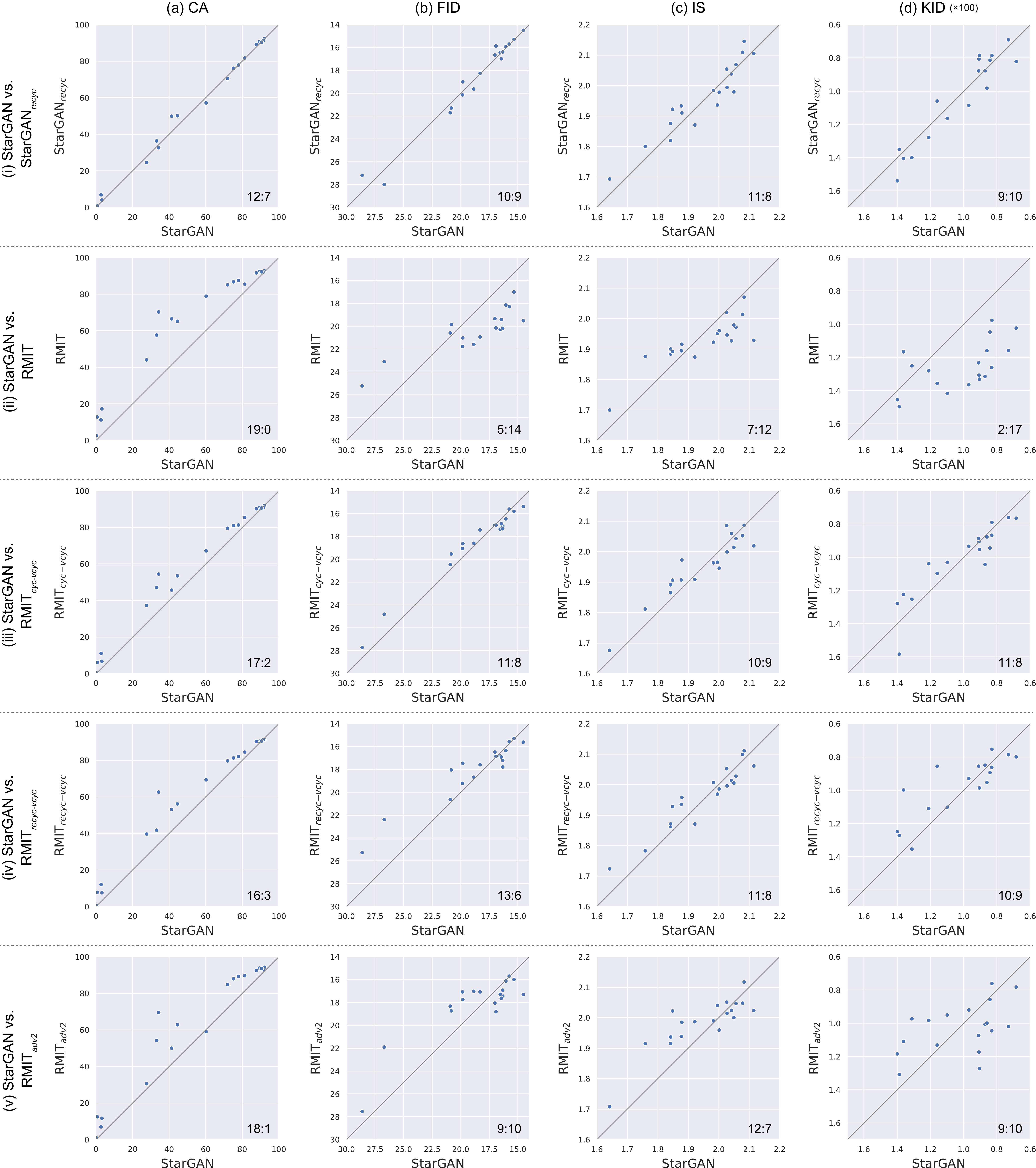}
  \caption{StarGAN vs. (i) StarGAN$_{recyc}$, (ii) RMIT, (iii) RMIT$_{cyc\mathchar`-vcyc}$, (iv) RMIT$_{recyc\mathchar`-vcyc}$, and (v) RMIT$_{adv2}$. For easy viewing, we exclude samples for which the FID is over 30 or the KID ($\times 100$) is over 3 when plotting. For the CA and IS (a larger value is better), the axis is in forward scale, while for the FID and KID (a smaller value is better) the axis is in inverse scale. In all the figures, the dots correspond to each condition listed in Table~\ref{tab:condition}. Dots above the diagonal line indicate that a compared model is better than the naive StarGAN. The number in the bottom-right corner of each figure indicates the win-to-loss ratio, i.e., how many times the compared model wins or loses against StarGAN. For the CA, RMIT and its advanced variants (RMIT$_{cyc\mathchar`-vcyc}$, RMIT$_{recyc\mathchar`-vcyc}$, RMIT$_{adv2}$) outperform StarGAN in a high ratio. For the FID, IS, and KID, RMIT achieves worse scores. However, this degradation is recovered by the advanced RMITs, and these are comparable with StarGAN.}
  \label{fig:vs}
  \vspace{-0mm}
\end{figure*}

We summarize the comparison between StarGAN and the other five models across all 19 conditions in Figure~\ref{fig:vs}. Regarding the CA, RMIT and the advanced RMITs outperform StarGAN in most cases. Even the worse case (i.e., RMIT$_{recyc\mathchar`-vcyc}$) outperforms StarGAN for $84.2\%$ ($= 16/19$) of conditions. Regarding the FID, RMIT achieves worse scores. However, this degradation is not reflected in the advanced RMITs. Even in the worse case (i.e., RMIT$_{recyc\mathchar`-vcyc}$), the win/lose rate is $47.4\%/52.6\%$. For the IS and KID, we observe similar tendencies to the FID. These results confirm the claims made at the beginning of this section. From the results, we conclude that the virtual cycle consistency with the advanced techniques provides a reasonable solution for label-noise robust multi-domain image-to-image translation.

\clearpage
\subsection{Analysis of evaluation metrics}
\label{subsec:analysis_eval}
As discussed in previous studies~\cite{LTheisICLR2016,MLucicNeurIPS2018,SBarrattICMLW2018}, the evaluation of GANs is challenging, partially owing to the lack of an explicit likelihood measure. Motivated by this fact, evaluation metrics have been keenly studied, and various evaluation metrics have been proposed. Among these, we utilized two evaluation metrics, the classification accuracy (CA)~\cite{YChoiCVPR2018,BZhaoECCV2018} and Fr\'{e}chet Inception distance (FID)~\cite{MHeuselNIPS2017}, which are typically used in multi-domain image-to-image translation or image generation. To validate this choice, we conducted an additional analysis on the evaluation metrics. In particular, we evaluated the models using two other popular metrics, i.e., the Inception score (IS)~\cite{TSalimansNIPS2016} and Kernel Inception distance (KID)~\cite{MBinkowskiICLR2018}. We then examined the correlations among the evaluation metrics to determine which choice of evaluation metrics is reasonable for a comprehensive analysis. In this section, we first describe the procedure for calculating the evaluation metrics and subsequently present the results.

\subsubsection{Calculation procedure}
\textbf{CA:} We used the CA to evaluate whether a translated image belongs to the correct target domain. We first trained a classifier (in particular, we used PreAct ResNet-18~\cite{KHeECCV2016}) using clean-labeled training data. We trained it independently of image translation models. We subsequently translated images in the test set to the domain that is different from the original domain using each image translation model. Finally, we calculated the accuracy of the translated images using the above-mentioned classifier. By this definition, when a nonconversion model is learned (such a model tends to be learned in a severely noisy case), the CA becomes close to $0 \%$. A larger CA is better.

\smallskip\noindent\textbf{FID:}
To evaluate the fidelity of the generated images, we used the FID, which measures the 2-Wasserstein distance between real data and generated data in the Inception embeddings~\cite{CSzegedyCVPR2016}. In particular, we first translated an image in the test set using the labels of another image. We subsequently calculated the FID between the translated samples and all the real samples in the training set. A smaller FID is better.

\smallskip\noindent\textbf{IS:}
The IS is calculated based on the KL-divergence between the conditional class distribution $p(y|{\bm x})$ and marginal class distribution $p(y) \approx \mathbb{E}_{{\bm x} \sim p^f({\bm x})}p(y|{\bm x})$. When $p(y|{\bm x})$ has a low entropy (i.e., images are classifiable) in addition to $p(y)$ having a high entropy (i.e., images have high divergence), the IS becomes high. To estimate $p(y|{\bm x})$ and $p(y)$, we utilized the Inception-v3~\cite{CSzegedyCVPR2016}. The original Inception-v3 was trained on the ImageNet dataset of which contents are different from the dataset used in this study. Therefore, following the previous studies~\cite{XHuangECCV2018,ARomeroArXiv2018}, we employed the Inception-v3 fine-tuned on the target dataset. We calculated the IS for the generated images used to calculate the FID. A larger IS is better.

\smallskip\noindent\textbf{KID:}
The KID measures the squared maximum mean discrepancy (MMD) between real data and generated data in the Inception embeddings~\cite{CSzegedyCVPR2016}. The advantage of the KID compared to the FID is that it has an unbiased estimator, unlike the FID. Similar to the IS, we calculated the KID for the generated images used to calculate the FID. To ensure consistency, we calculate the mean KID averaged over 10 different splits of size 50. A smaller KID is better.

\subsubsection{Correlation among evaluation metrics}
\begin{figure*}[h]
  \centering
  \includegraphics[width=0.7\textwidth]{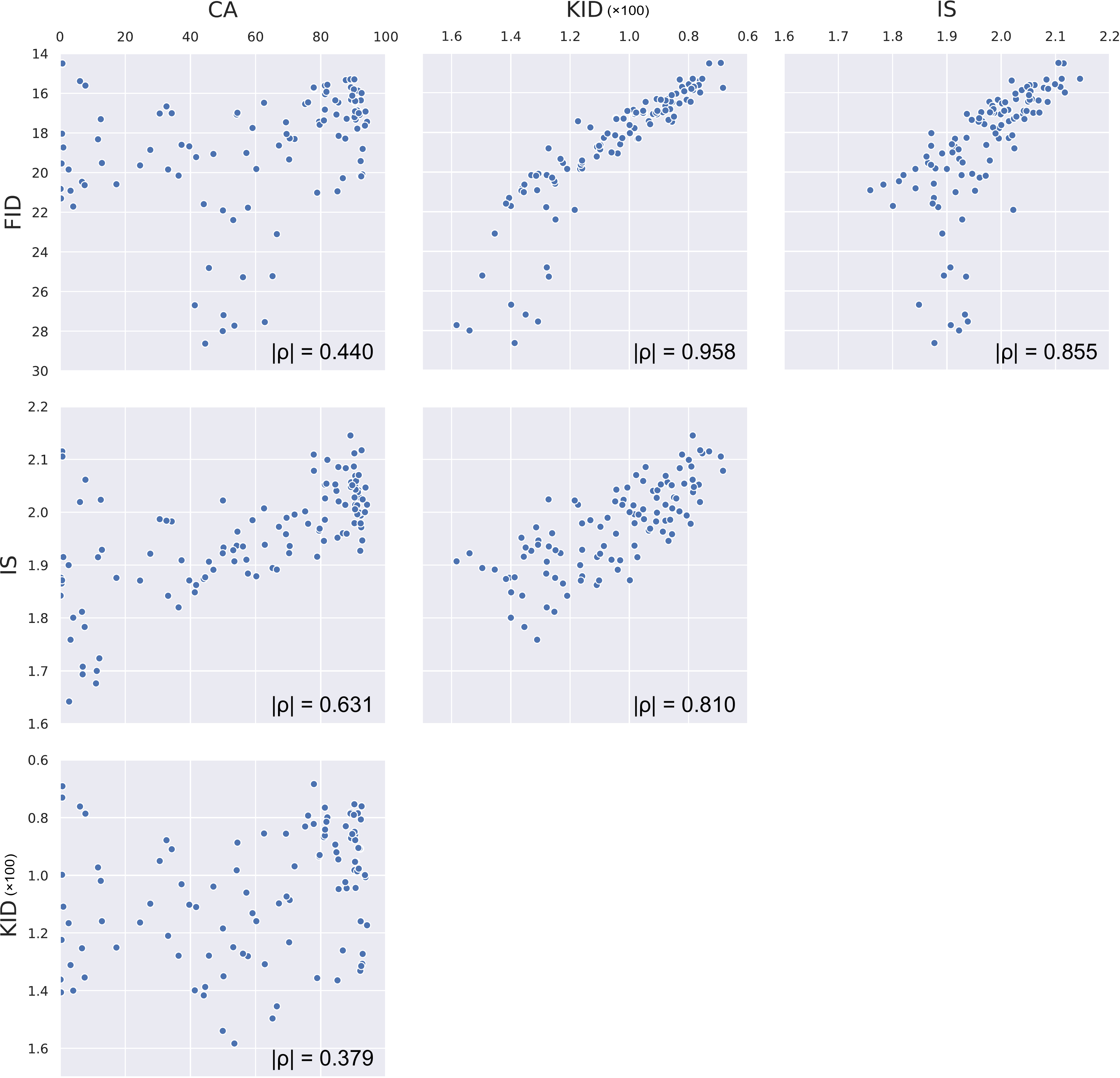}
  \caption{Correlation among evaluation metrics for all models and conditions. For easy viewing, we exclude samples for which the FID is over 30 or the KID ($\times 100$) is over 3 when plotting. For the CA and IS, a larger value is better, while for the FID and KID, a smaller value is better. The number in the bottom-right corner of each figure indicates the absolute value of the Spearman rank correlation, i.e., $| \rho |$. A higher $| \rho |$ value indicates a higher correlation in terms of the ranking. There is a high correlation between the FID and KID (0.958). In contrast, there are low correlations between the CA and FID (0.440) and between the CA and KID (0.379). This indicates that the CA and FID (or KID) are orthogonal, and the usage of these evaluation metrics is reasonable for analyzing a model in a comprehensive manner in our task.}
  \label{fig:corr}
  \vspace{-0mm}
\end{figure*}

\begin{figure*}[h]
  \centering
  \includegraphics[width=\textwidth]{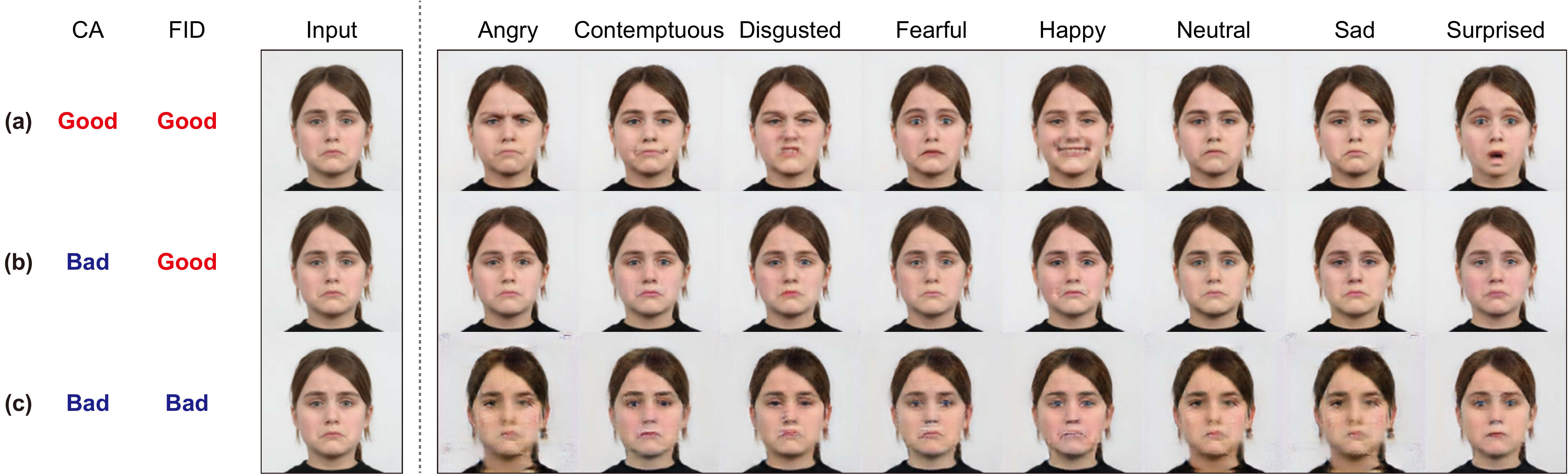}
  \caption{(Best zoomed in.) Generated images that achieve (a) a good CA and good FID (StarGAN with a naive classifier (no noise)), (b) a bad CA but good FID (StarGAN with forward correction (symmetric noise with $\mu = 0.75$)), and (c) a bad CA and bad FID (StarGAN with co-teaching (symmetric noise with $\mu = 0.75$)). In (b), a nonconvrsion model is learned. Such a model can achieve a good FID, but its CA is close to zero. In (c), the model not only fails to learn a meaningful conversion, but also results in blurring artifacts (particularly, in the ``angry'' row). Such a model degrades the FID as well as the CA.}
  \label{fig:ca_vs_fid}
  \vspace{-0mm}
\end{figure*}

We illustrate the correlation among evaluation metrics in Figure~\ref{fig:corr}. We summarize the results for all the models (the six models listed in Table~\ref{tab:model}) and all the conditions (the 19 conditions listed in Table~\ref{tab:condition}). Namely, we sum over $6 \:({\rm models}) \times 19 \:({\rm conditions}) = 114$ states. To numerically analyze the correlation, we calculate the absolute value of the Spearman rank correlation $|\rho|$. We present the result in the bottom-right corner of each figure. These results indicate that the FID and KID have a high correlation ($|\rho| = 0.958$), the CA and FID (or KID) have a low correlation ($|\rho| = 0.440$ (or $|\rho| = 0.379$)), and the IS has a comparatively high correlation with the FID ($|\rho| = 0.855$) and KID ($|\rho| = 0.810$) and a medium correlation with the CA ($|\rho| = 0.631$). These results confirm that evaluation using a combination of the CA and FID (or a combination of the CA and KID) is reasonable to comprehensively analyze the models in our task. As reference, we present the generated images that achieve (a) a good CA and good FID, (b) a bad CA but good FID, and (c) a bad CA and bad FID in Figure~\ref{fig:ca_vs_fid}. These evaluation metrics are orthogonal and it is possible to achieve a high performance in terms of either of them.

\subsubsection{Precise numerical values}
For reference, we report the precise numerical values of the CA, FID, IS and KID in Tables~\ref{tab:score_cls_no}--\ref{tab:score_tcls_asym}. These represent the extended versions of Tables~\ref{tab:score_cls} and \ref{tab:score_rcls} in the main text.

\clearpage
\begin{table*}[h]
  \centering
  \scalebox{0.8}{
    \begin{tabular}{l|cccc}
      \bhline{1pt}
      \multirow{3}{*}{Model}
      & \multicolumn{4}{c}{No noise}
      \\ \cline{2-5}
      & CA
      & FID
      & IS
      & KID
      \\
      &
      &
      &
      & \scriptsize{($\times 100$)}
      \\ \bhline{0.75pt}
      StarGAN
      & 92.1 & \textbf{16.3} & \textbf{2.03} & \textbf{0.91}
      \\
      StarGAN$_{recyc}$
      & 92.3 & \textbf{16.4} & 1.99 & \textbf{0.81}
      \\ \hline
      RMIT
      & \textbf{92.8} & 20.1 & 1.95 & 1.31
      \\
      RMIT$_{cyc\mathchar`-vcyc}$
      & 91.9 & 17.1 & 2.00 & \textbf{0.91}
      \\
      RMIT$_{recyc\mathchar`-vcyc}$
      & 91.2 & 17.8 & 2.00 & 0.98
      \\
      RMIT$_{adv2}$
      & \textbf{94.2} & 17.4 & \textbf{2.01} & 1.17
      \\ \bhline{1pt}
    \end{tabular}
  }
  \vspace{1mm}
  \caption{Quantitative results in the clean settings. This table is the extended version of Table~\ref{tab:score_cls}. A larger CA, smaller FID, larger IS, and smaller KID are better. We multiply the KID by a factor of 100. The two best scores are boldfaced.}
  \label{tab:score_cls_no}
\end{table*}

\begin{table*}[h]
  \centering
  \scalebox{0.8}{
    \begin{tabular}{l|cccc|cccc|cccc}
      \bhline{1pt}
      \multirow{3}{*}{Model}
      & \multicolumn{12}{c}{Naive classifier (symmetric noise)}
      \\ \cline{2-13}
      & \multicolumn{4}{c|}{0.25}
      & \multicolumn{4}{c|}{0.5}
      & \multicolumn{4}{c}{0.75}
      \\ \cline{2-13}
      & CA
      & FID
      & IS
      & KID
      & CA
      & FID
      & IS
      & KID
      & CA
      & FID
      & IS
      & KID
      \\
      &
      &
      &
      & \scriptsize{($\times 100$)}
      &
      &
      &
      & \scriptsize{($\times 100$)}
      &
      &
      &
      & \scriptsize{($\times 100$)}
      \\ \bhline{0.75pt}
      StarGAN
      & 33.2 & 19.8 & 1.84 & 1.21
      & 3.2  & 20.9 & 1.76 & 1.31
      & 0.2  & 20.8 & 1.84 & 1.36
      \\
      StarGAN$_{recyc}$
      & 36.4 & 20.2 & 1.82 & 1.28
      & 4.0  & 21.7 & 1.80 & 1.40
      & 0.3  & 21.3 & 1.88 & 1.41
      \\ \hline
      RMIT
      & \textbf{57.7} & 21.8 & 1.88 & 1.28
      & \textbf{17.3} & 20.6 & \textbf{1.88} & \textbf{1.25}
      & \textbf{2.6}  & 19.8 & \textbf{1.90} & 1.17
      \\
      RMIT$_{cyc\mathchar`-vcyc}$
      & 47.1 & \textbf{19.1} & \textbf{1.89} & \textbf{1.04}
      & 6.7  & \textbf{20.5} & 1.81 & \textbf{1.25}
      & 0.5  & 19.5 & 1.87 & 1.22
      \\
      RMIT$_{recyc\mathchar`-vcyc}$
      & 41.8 & 19.2 & 1.86 & 1.11
      & 7.6  & 20.6 & 1.78 & 1.35
      & 0.6  & \textbf{18.0} & 1.87 & \textbf{1.00}
      \\
      RMIT$_{adv2}$
      & \textbf{54.2} & \textbf{17.1} & \textbf{1.94} & \textbf{0.98}
      & \textbf{11.7} & \textbf{18.3} & \textbf{1.91} & \textbf{0.97}
      & \textbf{1.0}  & \textbf{18.7} & \textbf{1.92} & \textbf{1.11}
      \\ \bhline{1pt}
    \end{tabular}
  }
  \vspace{1mm}
  \caption{Quantitative results using models without a label-noise robust classifier in the symmetric noise settings. This table is the extended version of Table~\ref{tab:score_cls}. The second row indicates the noise rate $\mu$. A larger CA, smaller FID, larger IS, and smaller KID are better. We multiply the KID by a factor of 100. The two best scores are boldfaced.}
  \label{tab:score_cls_sym}
\end{table*}

\begin{table*}[h]
  \centering
  \scalebox{0.8}{
    \begin{tabular}{l|cccc|cccc|cccc}
      \bhline{1pt}
      \multirow{3}{*}{Model}
      & \multicolumn{12}{c}{Naive classifier (asymmetric noise)}
      \\ \cline{2-13}
      & \multicolumn{4}{c|}{0.15}
      & \multicolumn{4}{c|}{0.3}
      & \multicolumn{4}{c}{0.45}
      \\ \cline{2-13}
      & CA
      & FID
      & IS
      & KID
      & CA
      & FID
      & IS
      & KID
      & CA
      & FID
      & IS
      & KID
      \\
      &
      &
      &
      & \scriptsize{($\times 100$)}
      &
      &
      &
      & \scriptsize{($\times 100$)}
      &
      &
      &
      & \scriptsize{($\times 100$)}
      \\ \bhline{0.75pt}
      StarGAN
      & 72.0 & 18.3 & \textbf{2.00} & 0.97
      & 60.2 & 19.8 & 1.88 & 1.16
      & 27.7 & 18.9 & \textbf{1.92} & 1.10
      \\
      StarGAN$_{recyc}$
      & 70.5 & 18.3 & 1.94 & 1.09
      & 57.2 & 19.0 & 1.91 & \textbf{1.06}
      & 24.6 & 19.6 & 1.87 & 1.16
      \\ \hline
      RMIT
      & \textbf{85.2} & 20.9 & 1.95 & 1.36
      & \textbf{78.9} & 21.0 & 1.92 & 1.36
      & \textbf{44.1} & 21.6 & 1.87 & 1.42
      \\
      RMIT$_{cyc\mathchar`-vcyc}$
      & 79.5 & \textbf{17.4} & 1.97 & \textbf{0.93}
      & 67.2 & 18.6 & \textbf{1.97} & 1.10
      & 37.3 & \textbf{18.6} & 1.91 & \textbf{1.03}
      \\
      RMIT$_{recyc\mathchar`-vcyc}$
      & 79.7 & 17.6 & 1.97 & \textbf{0.93}
      & \textbf{69.4} & \textbf{17.5} & 1.96 & \textbf{0.86}
      & \textbf{39.7} & 18.7 & 1.87 & 1.10
      \\
      RMIT$_{adv2}$
      & \textbf{84.8} & \textbf{17.1} & \textbf{2.04} & \textbf{0.92}
      & 59.1 & \textbf{17.7} & \textbf{1.99} & 1.13
      & 30.6 & \textbf{17.0} & \textbf{1.99} & \textbf{0.95}
      \\ \bhline{1pt}
    \end{tabular}
  }
  \vspace{1mm}
  \caption{Quantitative results using models without a label-noise robust classifier in the asymmetric noise settings. This table is the extended version of Table~\ref{tab:score_cls}. The second row indicates the noise rate $\mu$. A larger CA, smaller FID, larger IS, and smaller KID are better. We multiply the KID by a factor of 100. The two best scores are boldfaced.}
  \label{tab:score_cls_asym}
\end{table*}

\begin{table*}[h]
  \centering
  \scalebox{0.8}{
    \begin{tabular}{l|cccc|cccc|cccc}
      \bhline{1pt}
      \multirow{3}{*}{Model}
      & \multicolumn{12}{c}{Forward correction (symmetric noise)}
      \\ \cline{2-13}
      & \multicolumn{4}{c|}{0.25}
      & \multicolumn{4}{c|}{0.5}
      & \multicolumn{4}{c}{0.75}
      \\ \cline{2-13}
      & CA
      & FID
      & IS
      & KID
      & CA
      & FID
      & IS
      & KID
      & CA
      & FID
      & IS
      & KID
      \\
      &
      &
      &
      & \scriptsize{($\times 100$)}
      &
      &
      &
      & \scriptsize{($\times 100$)}
      &
      &
      &
      & \scriptsize{($\times 100$)}
      \\ \bhline{0.75pt}
      StarGAN
      & 75.2 & \textbf{16.5} & \textbf{2.00} & \textbf{0.83}
      & 34.3 & 17.0 & 1.98 & 0.91
      & 0.7  & \textbf{14.5} & \textbf{2.12} & \textbf{0.73}
      \\
      StarGAN$_{recyc}$
      & 76.1 & \textbf{16.5} & 1.98 & \textbf{0.79}
      & 32.7 & \textbf{16.7} & 1.98 & \textbf{0.88}
      & 0.7  & \textbf{14.5} & \textbf{2.11} & \textbf{0.69}
      \\ \hline
      RMIT
      & \textbf{86.8} & 20.3 & 1.96 & 1.26
      & \textbf{70.3} & 19.3 & 1.92 & 1.23
      & \textbf{12.9} & 19.5 & 1.93 & 1.16
      \\
      RMIT$_{cyc\mathchar`-vcyc}$
      & 81.0 & 17.4 & 1.95 & 0.87
      & 54.4 & 17.0 & 1.96 & 0.89
      & 6.1  & 15.4 & 2.02 & 0.76
      \\
      RMIT$_{recyc\mathchar`-vcyc}$
      & 81.3 & 16.8 & \textbf{1.99} & 0.86
      & 62.6 & \textbf{16.5} & \textbf{2.01} & \textbf{0.86}
      & 7.8  & 15.6 & 2.06 & 0.79
      \\
      RMIT$_{adv2}$
      & \textbf{88.0} & 17.3 & 1.96 & 1.05
      & \textbf{69.5} & 18.1 & \textbf{1.99} & 1.07
      & \textbf{12.5} & 17.3 & 2.02 & 1.02
      \\ \bhline{1pt}
    \end{tabular}
  }
  \vspace{1mm}
  \caption{Quantitative results using models with forward correction in the symmetric noise settings. This table is the extended version of Table~\ref{tab:score_rcls}. The second row indicates the noise rate $\mu$. A larger CA, smaller FID, larger IS, and smaller KID are better. We multiply the KID by a factor of 100. The two best scores are boldfaced.}
  \label{tab:score_fcls_sym}
\end{table*}

\begin{table*}[h]
  \centering
  \scalebox{0.8}{
    \begin{tabular}{l|cccc|cccc|cccc}
      \bhline{1pt}
      \multirow{3}{*}{Model}
      & \multicolumn{12}{c}{Forward correction (asymmetric noise)}
      \\ \cline{2-13}
      & \multicolumn{4}{c|}{0.15}
      & \multicolumn{4}{c|}{0.3}
      & \multicolumn{4}{c}{0.45}
      \\ \cline{2-13}
      & CA
      & FID
      & IS
      & KID
      & CA
      & FID
      & IS
      & KID
      & CA
      & FID
      & IS
      & KID
      \\
      &
      &
      &
      & \scriptsize{($\times 100$)}
      &
      &
      &
      & \scriptsize{($\times 100$)}
      &
      &
      &
      & \scriptsize{($\times 100$)}
      \\ \bhline{0.75pt}
      StarGAN
      & 91.6 & \textbf{16.9} & \textbf{2.04} & \textbf{0.91}
      & 89.3 & \textbf{16.3} & \textbf{2.06} & \textbf{0.87}
      & 90.6 & \textbf{16.4} & \textbf{2.05} & \textbf{0.86}
      \\
      StarGAN$_{recyc}$
      & 91.4 & \textbf{15.9} & \textbf{2.04} & \textbf{0.78}
      & 90.6 & \textbf{16.4} & \textbf{2.07} & 0.88
      & 90.4 & 17.0 & 1.98 & 0.98
      \\ \hline
      RMIT
      & \textbf{92.2} & 20.2 & 1.93 & 1.33
      & \textbf{92.4} & 20.2 & 1.97 & 1.31
      & \textbf{92.3} & 19.4 & 1.98 & 1.16
      \\
      RMIT$_{cyc\mathchar`-vcyc}$
      & 90.9 & 17.0 & \textbf{2.06} & 0.95
      & 90.7 & 17.3 & 2.04 & 1.04
      & 90.6 & \textbf{16.9} & \textbf{2.01} & \textbf{0.88}
      \\
      RMIT$_{recyc\mathchar`-vcyc}$
      & 91.2 & \textbf{16.9} & 2.01 & 0.99
      & 90.4 & 17.2 & 2.03 & \textbf{0.85}
      & 90.5 & \textbf{16.9} & \textbf{2.01} & 0.95
      \\
      RMIT$_{adv2}$
      & \textbf{92.9} & 18.8 & 2.02 & 1.27
      & \textbf{93.8} & 16.9 & 2.05 & 1.01
      & \textbf{93.6} & 17.6 & 2.00 & 1.00
      \\ \bhline{1pt}
    \end{tabular}
  }
  \vspace{1mm}
  \caption{Quantitative results using models with forward correction in the asymmetric noise settings. This table is the extended version of Table~\ref{tab:score_rcls}. The second row indicates the noise rate $\mu$. A larger CA, smaller FID, larger IS, and smaller KID are better. We multiply the KID by a factor of 100. The two best scores are boldfaced.}
  \label{tab:score_fcls_asym}
\end{table*}

\begin{table*}[h]
  \centering
  \scalebox{0.8}{
    \begin{tabular}{l|cccc|cccc|cccc}
      \bhline{1pt}
      \multirow{3}{*}{Model}
      & \multicolumn{12}{c}{Co-teaching (symmetric noise)}
      \\ \cline{2-13}
      & \multicolumn{4}{c|}{0.25}
      & \multicolumn{4}{c|}{0.5}
      & \multicolumn{4}{c}{0.75}
      \\ \cline{2-13}
      & CA
      & FID
      & IS
      & KID
      & CA
      & FID
      & IS
      & KID
      & CA
      & FID
      & IS
      & KID
      \\
      &
      &
      &
      & \scriptsize{($\times 100$)}
      &
      &
      &
      & \scriptsize{($\times 100$)}
      &
      &
      &
      & \scriptsize{($\times 100$)}
      \\ \bhline{0.75pt}
      StarGAN
      & 77.9 & 15.8 & 2.08 & \textbf{0.68}
      & 44.6 & 28.6 & 1.88 & 1.39
      & 2.8  & 46.8 & 1.64 & 3.51
      \\
      StarGAN$_{recyc}$
      & 77.9 & 15.7 & 2.11 & 0.82
      & 50.2 & 27.2 & 1.93 & 1.35
      & 6.9  & 49.8 & 1.69 & 3.94
      \\ \hline
      RMIT
      & \textbf{87.6} & 18.3 & \textbf{2.01} & 1.02
      & \textbf{65.2} & \textbf{25.2} & 1.89 & 1.50
      & \textbf{11.3} & 44.9 & 1.70 & 3.62
      \\
      RMIT$_{cyc\mathchar`-vcyc}$
      & 81.3 & \textbf{15.6} & \textbf{2.05} & \textbf{0.77}
      & 53.5 & 27.7 & 1.91 & 1.58
      & 11.0 & \textbf{39.8} & 1.68 & \textbf{3.16}
      \\
      RMIT$_{recyc\mathchar`-vcyc}$
      & 82.1 & \textbf{15.6} & 2.10 & 0.80
      & 56.1 & \textbf{25.3} & \textbf{1.94} & \textbf{1.27}
      & \textbf{12.1} & \textbf{44.0} & \textbf{1.72} & \textbf{3.36}
      \\
      RMIT$_{adv2}$
      & \textbf{89.3} & 15.7 & \textbf{2.05} & 0.78
      & \textbf{62.9} & 27.5 & \textbf{1.94} & \textbf{1.31}
      & 7.0  & 55.8 & \textbf{1.71} & 4.54
      \\ \bhline{1pt}
    \end{tabular}
  }
  \vspace{1mm}
  \caption{Quantitative results using models with co-teaching in the symmetric noise settings. This table is the extended version of Table~\ref{tab:score_rcls}. The second row indicates the noise rate $\mu$. A larger CA, smaller FID, larger IS, and smaller KID are better. We multiply the KID by a factor of 100. The two best scores are boldfaced.}
  \label{tab:score_tcls_sym}
\end{table*}

\begin{table*}[h]
  \centering
  \scalebox{0.8}{
    \begin{tabular}{l|cccc|cccc|cccc}
      \bhline{1pt}
      \multirow{3}{*}{Model}
      & \multicolumn{12}{c}{Co-teaching (asymmetric noise)}
      \\ \cline{2-13}
      & \multicolumn{4}{c|}{0.15}
      & \multicolumn{4}{c|}{0.3}
      & \multicolumn{4}{c}{0.45}
      \\ \cline{2-13}
      & CA
      & FID
      & IS
      & KID
      & CA
      & FID
      & IS
      & KID
      & CA
      & FID
      & IS
      & KID
      \\
      &
      &
      &
      & \scriptsize{($\times 100$)}
      &
      &
      &
      & \scriptsize{($\times 100$)}
      &
      &
      &
      & \scriptsize{($\times 100$)}
      \\ \bhline{0.75pt}
      StarGAN
      & 87.7 & \textbf{15.3} & 2.08 & 0.83
      & 81.4 & \textbf{16.0} & 2.03 & \textbf{0.84}
      & 41.4 & 26.7 & 1.85 & 1.40
      \\
      StarGAN$_{recyc}$
      & 89.1 & \textbf{15.3} & \textbf{2.15} & 0.79
      & 81.8 & \textbf{15.9} & \textbf{2.05} & \textbf{0.81}
      & 49.9 & 28.0 & 1.92 & 1.54
      \\ \hline
      RMIT
      & \textbf{91.7} & 17.0 & 2.07 & 0.98
      & \textbf{85.5} & 18.1 & 2.02 & 1.05
      & \textbf{66.5} & 23.1 & 1.89 & 1.45
      \\
      RMIT$_{cyc\mathchar`-vcyc}$
      & 90.2 & 15.8 & 2.09 & 0.79
      & 85.4 & 16.5 & \textbf{2.09} & 0.95
      & 45.7 & 24.8 & 1.91 & 1.28
      \\
      RMIT$_{recyc\mathchar`-vcyc}$
      & 90.3 & \textbf{15.3} & 2.11 & \textbf{0.75}
      & 84.5 & 16.4 & \textbf{2.05} & 0.89
      & \textbf{53.2} & \textbf{22.4} & \textbf{1.93} & \textbf{1.25}
      \\
      RMIT$_{adv2}$
      & \textbf{92.6} & 16.0 & \textbf{2.12} & \textbf{0.76}
      & \textbf{89.7} & 16.1 & \textbf{2.05} & 0.86
      & 50.0 & \textbf{21.9} & \textbf{2.02} & \textbf{1.18}
      \\ \bhline{1pt}
    \end{tabular}
  }
  \vspace{1mm}
  \caption{Quantitative results using models with co-teaching in the asymmetric noise settings. This table is the extended version of Table~\ref{tab:score_rcls}. The second row indicates the noise rate $\mu$. A larger CA, smaller FID, larger IS, and smaller KID are better. We multiply the KID by a factor of 100. The two best scores are boldfaced.}
  \label{tab:score_tcls_asym}
\end{table*}

\clearpage
\subsection{Score trajectories}
\label{subsec:score_traj}
\begin{figure*}[h]
  \centering
  \vspace{-2mm}
  \includegraphics[width=0.92\textwidth]{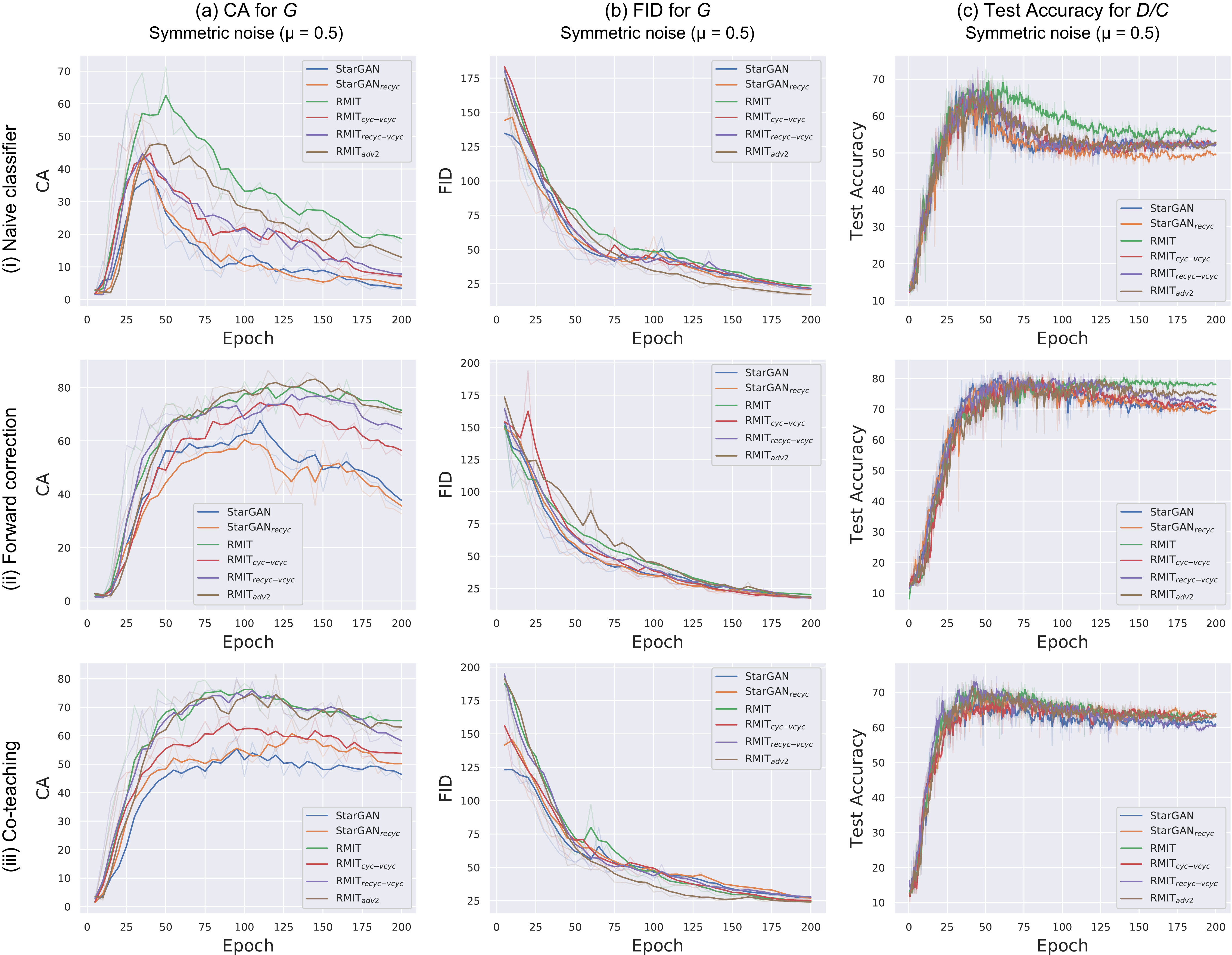}
  \caption{Score trajectories during the training in the symmetric noise setting ($\mu = 0.5$). (a) CA against number of epochs. This score was calculated for images generated by $G$. We computed the CA every five epochs. A larger CA is better. (b) FID against number of epochs. This score was also calculated for images generated by $G$. We computed the FID every five epochs. A smaller FID is better. (c) Test accuracy against number of epochs. This score was calculated for real images in the test set using the classifier in $D$/$C$. We computed the test accuracy every 100 iterations. A larger test accuracy is better. In all the figures, we smooth the graph for easy viewing.}
  \label{fig:traj_sym}
  \vspace{-0mm}
\end{figure*}

\begin{figure*}[h]
  \centering
  \includegraphics[width=0.92\textwidth]{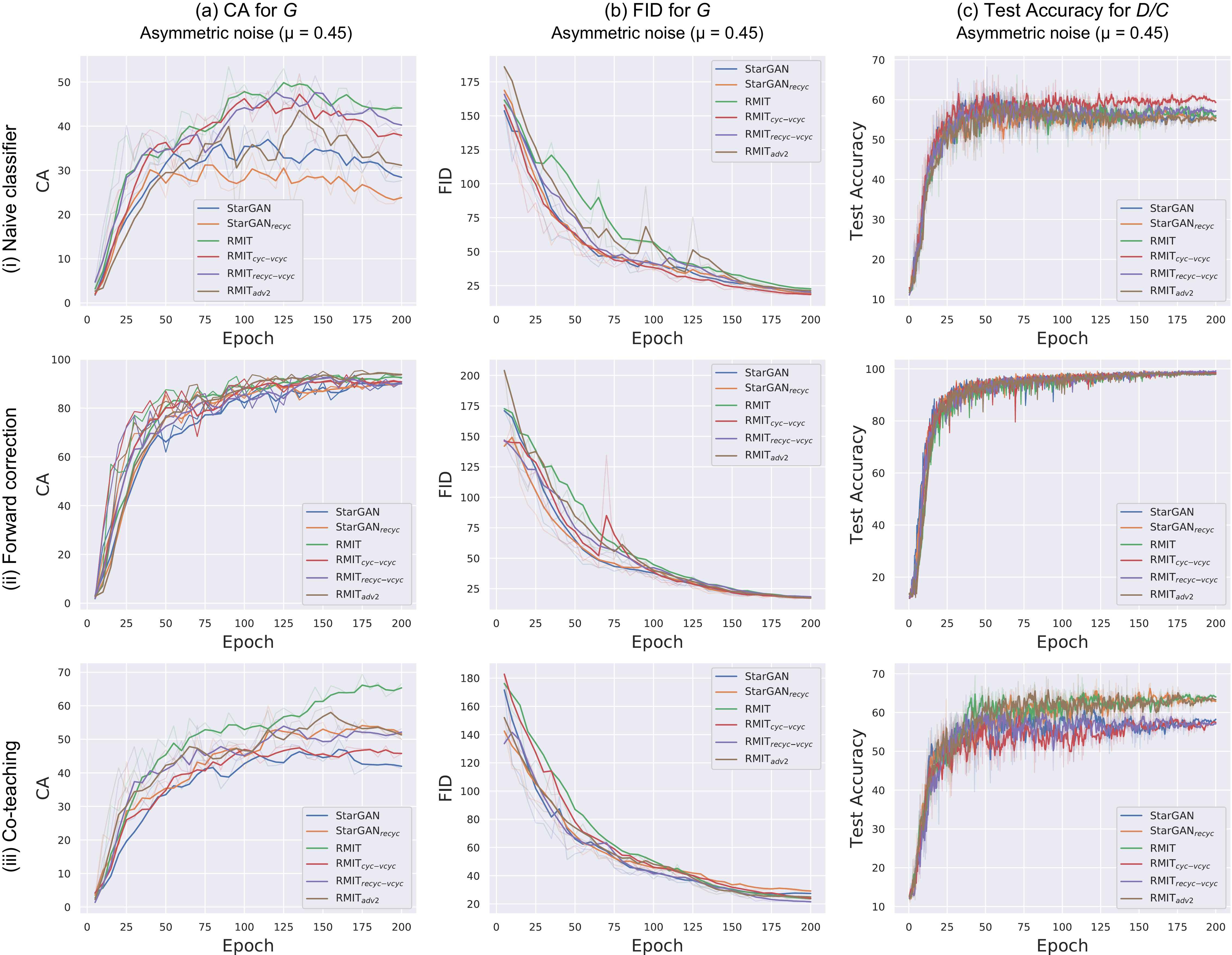}
  \caption{Score trajectories during the training in the asymmetric noise setting ($\mu = 0.45$). The view of the figure is the same as for Figure~\ref{fig:traj_sym}.}
  \label{fig:traj_asym}
  \vspace{-0mm}
\end{figure*}

We depict the score trajectories during the training in Figures~\ref{fig:traj_sym} and \ref{fig:traj_asym}. Figures~\ref{fig:traj_sym} and \ref{fig:traj_asym} show those in the symmetric ($\mu = 0.5$) and asymmetric ($\mu = 0.45$) noise settings, respectively. We plot three type scores: \textbf{(a) CA for $G$:} We calculated the CA for images generated by $G$. \textbf{(b) FID for $G$:} We calculated the FID for images generated by $G$. \textbf{(c) Test accuracy for $D/C$:} We calculated the classification accuracy for real images in the test set using the classifier in $D/C$.

In Figures~\ref{fig:traj_sym}(a) and \ref{fig:traj_asym}(a), RMIT and its advanced variants (RMIT$_{cyc\mathchar`-vcyc}$, RMIT$_{recyc\mathchar`-vcyc}$, and RMIT$_{adv2}$) consistently achieve a better CA across training than StarGAN and StarGAN$_{recyc}$, regardless of the type of classifier. These results confirm that the virtual cycle consistency loss is useful for improving the CA across training.

In Figure~\ref{fig:traj_sym}(i)(c), all the models achieve the best test accuracy in the early stage of training, and the scores decrease at the end of training. This is caused by the memorization effect~\cite{DArpitICML2017}, i.e., a DNN classifier first memorizes a simple pattern (i.e., clean labeled data), and then fits the noisy labeled data. Affected by these classifiers, a similar tendency is observed  in Figure~\ref{fig:traj_sym}(i)(a). These results indicate that it is important to consider the memorization effect when optimizing $G$ for the classifier in $D/C$. In Figure~\ref{fig:traj_sym}, this memorization effect is mitigated by two methods. The first is using label-noise robust classifiers (i.e., forward correction in Figure~\ref{fig:traj_sym}(ii) or co-teaching in Figure~\ref{fig:traj_sym}(iii)). The second is introducing the virtual cycle consistency loss (i.e., utilizing RMIT or its advanced variants). Another possible solution is early stopping. However, this is not practical for two reasons. First, early stopping requires the availability of clean validation data, which is not necessarily easy to collect in a practical setting. Second, as shown in Figure~\ref{fig:traj_sym}(b), the FID continues to improve across training, even when the CA degrades. Hence, early stopping results in a poor performance in terms of the FID.

\clearpage
\subsection{Effect of mixture rate}
\label{subsec:mixture}
\begin{figure*}[h]
  \centering
  \includegraphics[width=1\textwidth]{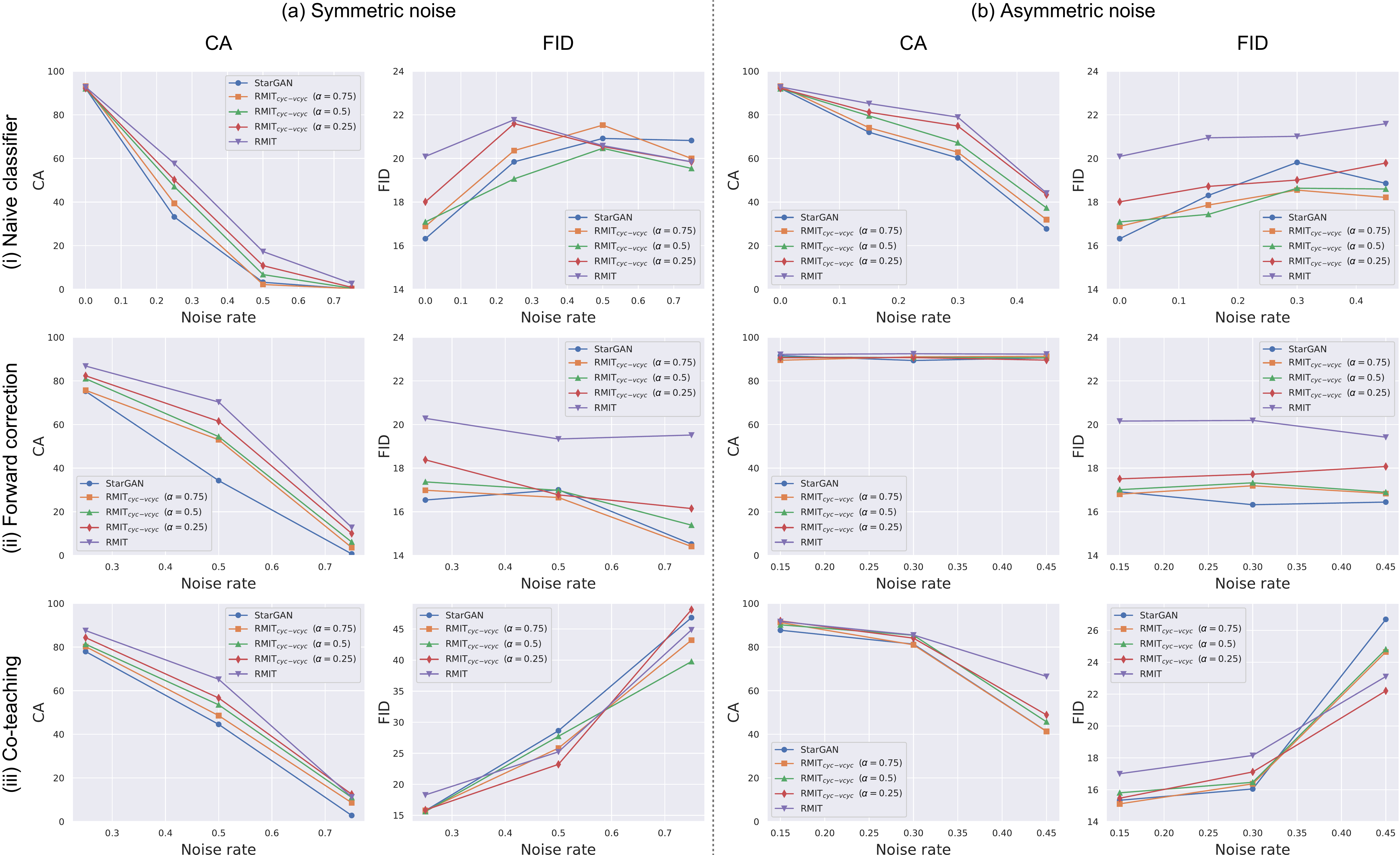}
  \caption{Comparison among the models with different mixture rates. In (a) and (b), we show the results in the symmetric and asymmetric noise settings, respectively. A larger CA is better, while a smaller FID is better.}
  \label{fig:mix}
  \vspace{-0mm}
\end{figure*}

In the main text, we fix the mixture rate $\alpha$ to 0.5 in RMIT$_{cyc\mathchar`-vcyc}$. To explore how the mixture rate affects the performance, we conducted the comparative experiments on the models with the different mixture rates. Figures~\ref{fig:mix}(a) and (b) show the results in the symmetric and asymmetric noise settings, respectively. In these figures, we observe that the CA decreases as the mixture rate increases regardless of the noise setting and the noise rate. These results indicate that we should decrease the mixture rate (i.e., weigh the virtual consistency loss highly) to improve the performance in terms of the CA.

Regarding the FID, a better or worse performance is case dependent and the score is not necessarily proportionate to the mixture rate and the noise rate. As shown in Figure~\ref{fig:ca_vs_fid}, multiple-type models (particularly, both a clean-label conditional model and a nonconversion model) can achieve a high performance in terms of the FID. In the noisy label setting, a model struggles among such various states. We argue that these various possibilities results in the nonmonotonic change.

Meanwhile, in some cases ((a-ii), (b-i), and (b-ii)), only the naive RMIT exhibits poor scores. As discussed in Section~\ref{subsec:advance}, this is possibly because the virtual cycle consistency loss is calculated between the generated images and is weak compared to the cycle consistency loss stemming from real images. However, this degradation is mitigated by a large margin by incorporating the cycle consistency loss (i.e., using RMIT$_{cyc\mathchar`-vcyc}$). This effect is observed even when the mixture rate is comparatively small ($\alpha = 0.25$). From these results, we confirm that incorporating the cycle consistency loss is a reasonable solution for mitigating the degradation in the virtual cycle consistency loss.

\clearpage
\section{Additional implementation details}
\label{sec:add_implementation}
\subsection{Implementation details on Section~\ref{subsec:comprehensive}}
\label{subsec:detail_comprehensive}
\noindent\textbf{Network architectures.}
We implemented the models based on the source code provided by the authors of StarGAN.\footnote{\url{https://github.com/yunjey/StarGAN}} The basic network architecture is the same as that utilized in the StarGAN study~\cite{YChoiCVPR2018}. The generator network is composed of downsampling, residual~\cite{KHeCVPR2016}, and upsampling layers, as well as incorporating instance normalization~\cite{DUlyanovArXiv2016}. The discriminator network is configured as PatchGAN~\cite{CLiECCV2016}. We list the details of network architectures in Table~\ref{tab:arch_128}. In the table, \textit{Conv} and \textit{Deconv} indicate convolutional and deconvolutional (i.e., fractionally strided convolutional) layers, respectively, and \textit{ReLU} and \textit{LReLU} denote rectified linear~\cite{VNairICML2010} and linear rectified linear~\cite{AMaasICML2013,BXuICMLW2015} units, respectively. In LReLU, we set the negative slope to 0.01. Furthermore, \textit{ResBlock} indicates a residual block~\cite{KHeCVPR2016}, and \textit{IN} is an abbreviation of instance normalization~\cite{DUlyanovArXiv2016}. In the original StarGAN~\cite{YChoiCVPR2018}, WGAN-GP~\cite{IGulrajaniNIPS2017} was used as a GAN objective. However, in this study, we replaced this with CT-GAN~\cite{XWeiICLR2018}, which is an improved version of WGAN-GP, to boost the performance. Note that in the experiments, we employed the CT-GAN regardless of the model. Therefore, all the models, including StarGAN and RMIT, obtain the benefits equally. Owing to this modification, we added a dropout to $D$/$C$, as listed in Table~\ref{tab:arch_128}(b).

\smallskip\noindent\textbf{Training settings.}
As discussed in Section~\ref{subsubsec:experimental_setup}, it is impractical or laborious to tune the training parameters depending on a label-noise setting when clean labels are not available. Therefore, we trained the models using standard parameters, which are typically employed in a clean-label setting. Namely, we used the same parameters as in the StarGAN study~\cite{YChoiCVPR2018}. More precisely, we set the trade-off parameters to $\lambda_{cls} = 1$ and $\lambda_{cyc} = 10$. We trained the models using the Adam optimizer~\cite{DPKingmaICLR2015} with $\beta_1 = 0.5$ and $\beta_2 = 0.999$. Furthermore, we set the learning rate to 0.0001 for the first 100 epochs, and linearly decreased this to 0 over the next 100 epochs. We updated $D$/$C$ five times per update of $G$, and the batch size was set to 16. For data augmentation, we flip images horizontally with a probability of 0.5.

\begin{table}[h]
  \centering  
  \begin{tabular}{cc}
    \begin{minipage}[t]{0.48\linewidth}
      \scalebox{0.8}{
        \begin{tabular}{l|l}        
          \bhline{1pt}
          Layer & Output shape
          \\ \bhline{0.75pt}
          Input: ${\bm x} \in \mathbb{R}^{128 \times 128 \times 3}$ and
          $y \in \{ 1, \dots, c \}$
                & $128 \times 128 \times (3 + c)$
          \\ \hline
          $7 \times 7$, stride=1 Conv 64, IN, ReLU
                & $128 \times 128 \times 64$
          \\
          $4 \times 4$, stride=2 Conv 128, IN, ReLU
                & $64 \times 64 \times 128$
          \\
          $4 \times 4$, stride=2 Conv 256, IN, ReLU
                & $32 \times 32 \times 256$
          \\ \hline
          $[3 \times 3] \times 2$ ResBlock
                & $32 \times 32 \times 256$
          \\
          $[3 \times 3] \times 2$ ResBlock
                & $32 \times 32 \times 256$
          \\
          $[3 \times 3] \times 2$ ResBlock
                & $32 \times 32 \times 256$
          \\
          $[3 \times 3] \times 2$ ResBlock
                & $32 \times 32 \times 256$
          \\
          $[3 \times 3] \times 2$ ResBlock
                & $32 \times 32 \times 256$
          \\
          $[3 \times 3] \times 2$ ResBlock
                & $32 \times 32 \times 256$
          \\ \hline
          $4 \times 4$, stride=2 Deconv 128, IN, ReLU
                & $64 \times 64 \times 128$
          \\
          $4 \times 4$, stride=2 Deconv 64, IN, ReLU
                & $128 \times 128 \times 64$
          \\ \hline
          $7 \times 7$, stride=1 Conv 3, Tanh $\rightarrow$ ${\bm x}'$
                & $128 \times 128 \times 3$
          \\ \bhline{1pt}
          \multicolumn{2}{c}{(a) Generator $G$}
          \\
      \end{tabular}
      }
    \end{minipage}

    \hfill{\hspace{3mm}}

    \begin{minipage}[t]{0.48\linewidth}
      \scalebox{0.8}{
        \begin{tabular}{l|l}
          \bhline{1pt}
          Layer & Output shape
          \\ \bhline{0.75pt}
          Input: ${\bm x} \in \mathbb{R}^{128 \times 128 \times 3}$
                & $128 \times 128 \times 3$
          \\ \hline
          $4 \times 4$, stride=2 Conv 64, LReLU
                & $64 \times 64 \times 64$
          \\
          $4 \times 4$, stride=2 Conv 128, LReLU, 0.2 Dropout
                & $32 \times 32 \times 128$
          \\
          $4 \times 4$, stride=2 Conv 256, LReLU, 0.2 Dropout
                & $16 \times 16 \times 256$
          \\
          $4 \times 4$, stride=2 Conv 512, LReLU, 0.2 Dropout
                & $8 \times 8 \times 512$
          \\
          $4 \times 4$, stride=2 Conv 1024, LReLU, 0.5 Dropout
                & $4 \times 4 \times 1024$
          \\
          $4 \times 4$, stride=2 Conv 2048, LReLU, 0.5 Dropout
                & $2 \times 2 \times 2048$
          \\ \hline
          $3 \times 3$, stride=1 Conv 1 for $D$
                & $2 \times 2 \times 1$
          \\
          $2 \times 2$, stride=1 Conv $c$ (zero pad) for $C$
                & $1 \times 1 \times c$
          \\ \bhline{1pt}
          \multicolumn{2}{c}{(b) Discriminator/classifier $D$/$C$}
          \\
        \end{tabular}
      }
    \end{minipage}
  \end{tabular}
  \vspace{1mm}
  \caption{Generator and discriminator/classifier network architectures
    utilized in Sections~\ref{subsec:comprehensive} and \ref{subsec:multiple}.}
  \label{tab:arch_128}
\end{table}

\subsection{Implementation details on Section~\ref{subsec:multiple}}
\label{subsec:detail_multi}
\noindent\textbf{Network architectures.}
We employed the same network architectures as described in Section~\ref{subsec:detail_comprehensive}.

\smallskip\noindent\textbf{Training settings.}
The training settings are almost the same as those described in Section~\ref{subsec:detail_comprehensive}. The difference lies in the number of epochs. We trained the models with the learning rate of 0.0001 for the first 10 epochs, and then linearly decreased this to 0 over the next 10 epochs. These settings are the same as in the StarGAN study~\cite{YChoiCVPR2018}.

\subsection{Implementation details on Section~\ref{subsec:real}}
\label{subsec:detail_real}
\noindent\textbf{Network architectures.}
As discussed in Section~\ref{subsec:real}, in the pre-experiments, we observed that a standard network and training setting does not perform well in the FER dataset~\cite{IGoodfellowNN2015}, possibly because this dataset is gray and not well aligned. However, we found that an identity mapping loss~\cite{YTaigmanICLR2017,JYZhuICCV2017} and attention mechanisms~\cite{APumarolaECCV2018} are useful for mitigating this problem. In particular, we replaced the last convolutional layer in $G$ with two parallel convolutional layers: One is used to calculate the color mask ${\bm x}_{color}$ and the other is used to define the attention mask ${\bm m}$. The final output is computed by ${\bm x}' = {\bm m} \cdot {\bm x}_{color} + (1 - {\bm m}) \cdot {\bm x}$, where ${\bm x}$ is the input image. For the discriminator, we used the residual network-based model~\cite{KHeCVPR2016} to further improve the performance. We designed its network architecture while referring to the state-of-the-art GAN studies~\cite{IGulrajaniNIPS2017,XWeiICLR2018,TMiyatoICLR2018b}. We list the details of network architectures in Table~\ref{tab:arch_48}.

\smallskip\noindent\textbf{Training settings.}
We used almost the same training settings as described in Section~\ref{subsec:detail_comprehensive}. The differences lie in the number of epochs and the introduction of the identity mapping loss. We trained the models with the learning rate of 0.0001 for the first 25 epochs, and then linearly decreased this to 0 over the next 25 epochs. We set the trade-off parameter $\lambda_{id}$, which weighs the importance of the identity mapping loss compared to the adversarial loss, to 5.

\begin{table}[h]
  \centering
  \begin{tabular}{cc}
    \begin{minipage}[t]{0.48\linewidth}
      \scalebox{0.8}{
        \begin{tabular}{l|l}
          \bhline{1pt}
          Layer & Output shape
          \\ \bhline{0.75pt}
          Input: ${\bm x} \in \mathbb{R}^{48 \times 48 \times 1}$ and
          $y \in \{ 1, \dots, c \}$
                & $48 \times 48 \times (1 + c)$
          \\ \hline
          $3 \times 3$, stride=1 Conv 64, IN, ReLU
                & $48 \times 48 \times 64$
          \\
          $4 \times 4$, stride=2 Conv 128, IN, ReLU
                & $24 \times 24 \times 128$
          \\
          $4 \times 4$, stride=2 Conv 256, IN, ReLU
                & $12 \times 12 \times 256$
          \\ \hline
          $[3 \times 3] \times 2$ ResBlock
                & $12 \times 12 \times 256$
          \\
          $[3 \times 3] \times 2$ ResBlock
                & $12 \times 12 \times 256$
          \\
          $[3 \times 3] \times 2$ ResBlock
                & $12 \times 12 \times 256$
          \\
          $[3 \times 3] \times 2$ ResBlock
                & $12 \times 12 \times 256$
          \\
          $[3 \times 3] \times 2$ ResBlock
                & $12 \times 12 \times 256$
          \\
          $[3 \times 3] \times 2$ ResBlock
                & $12 \times 12 \times 256$
          \\ \hline
          $4 \times 4$, stride=2 Deconv 128, IN, ReLU
                & $24 \times 24 \times 128$
          \\
          $4 \times 4$, stride=2 Deconv 64, IN, ReLU
                & $48 \times 48 \times 64$
          \\ \hline
          $3 \times 3$, stride=1 Conv 1, Tanh $\rightarrow$ ${\bm x}_{color}$
                & $48 \times 48 \times 1$
          \\
          $3 \times 3$, stride=1 Conv 1, Sigmoid $\rightarrow$ ${\bm m}$
                & $48 \times 48 \times 1$
          \\ \hline
          ${\bm m} \cdot {\bm x}_{color} + (1 - {\bm m}) \cdot {\bm x}$
                & $48 \times 48 \times 1$
          \\ \bhline{1pt}
          \multicolumn{2}{c}{(a) Generator $G$}
          \\
        \end{tabular}
      }
    \end{minipage}

    \hfill{\hspace{3mm}}

    \begin{minipage}[t]{0.48\linewidth}
      \scalebox{0.8}{
        \begin{tabular}{l|l}
          \bhline{1pt}
          Layer & Output shape
          \\ \bhline{0.75pt}
          Input: ${\bm x} \in \mathbb{R}^{48 \times 48 \times 1}$
                & $48 \times 48 \times 1$
          \\ \hline
          $[3 \times 3] \times 2$ ResBlock down
                & $24 \times 24 \times 64$
          \\
          $[3 \times 3] \times 2$ ResBlock down, 0.2 Dropout
                & $12 \times 12 \times 128$
          \\
          $[3 \times 3] \times 2$ ResBlock down, 0.5 Dropout
                & $6 \times 6 \times 256$
          \\
          $[3 \times 3] \times 2$ ResBlock down, 0.5 Dropout
                & $3 \times 3 \times 512$
          \\ \hline
          Global mean pooling
                & $1 \times 1 \times 512$
          \\ \hline
          $1 \times 1$, stride=1 Conv 1 for $D$
                & $1 \times 1 \times 1$
          \\
          $1 \times 1$, stride=1 Conv $c$ for $C$
                & $1 \times 1 \times c$
          \\ \bhline{1pt}
          \multicolumn{2}{c}{(b) Discriminator/classifier $D$/$C$}
          \\
        \end{tabular}
      }
    \end{minipage}
  \end{tabular}
  \vspace{1mm}
  \caption{Generator and discriminator/classifier network architectures utilized in Section~\ref{subsec:real}. ${\bm x}_{color}$ and ${\bm m}$ represent the color mask and attention mask, respectively.}
  \label{tab:arch_48}
\end{table}

\clearpage
\section{Additional generated images}
\label{sec:add_images}
\subsection{Extended results of Figure~\ref{fig:concept}}
\label{subsec:gen_concept_ex}

\begin{figure*}[h]
  \centering
  \includegraphics[width=\textwidth]{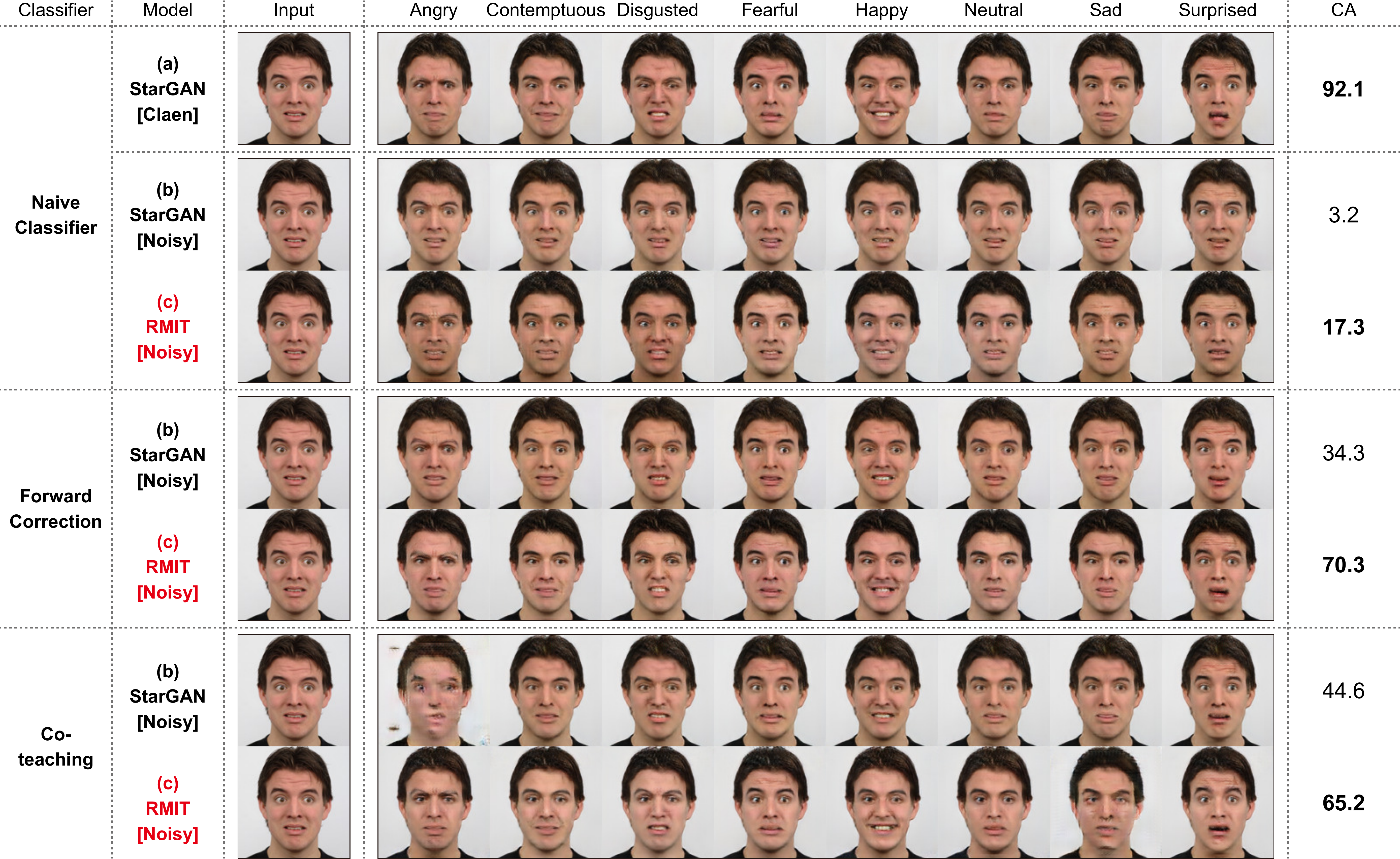}
  \caption{(Best zoomed in.) Examples of label-noise robust multi-domain image-to-image translation. This represents the extended results of Figure~\ref{fig:concept}. In (b), StarGAN in the noisy label setting (symmetric noise with $\mu = 0.5$) learns a nonconversion model when trained with a naive classifier (the second row). Even when trained with label-noise robust classifiers (the fourth and sixth rows), the translation is limited to partial changes while comparing to RMIT in (c). Indeed, in all the cases, StarGAN in the noisy label setting (b) achieves a lower CA than RMIT (c). We show the results for another person in Figure~\ref{fig:concept_ex2}.}
  \label{fig:concept_ex1}
\end{figure*}

\begin{figure*}[h]
  \centering
  \includegraphics[width=\textwidth]{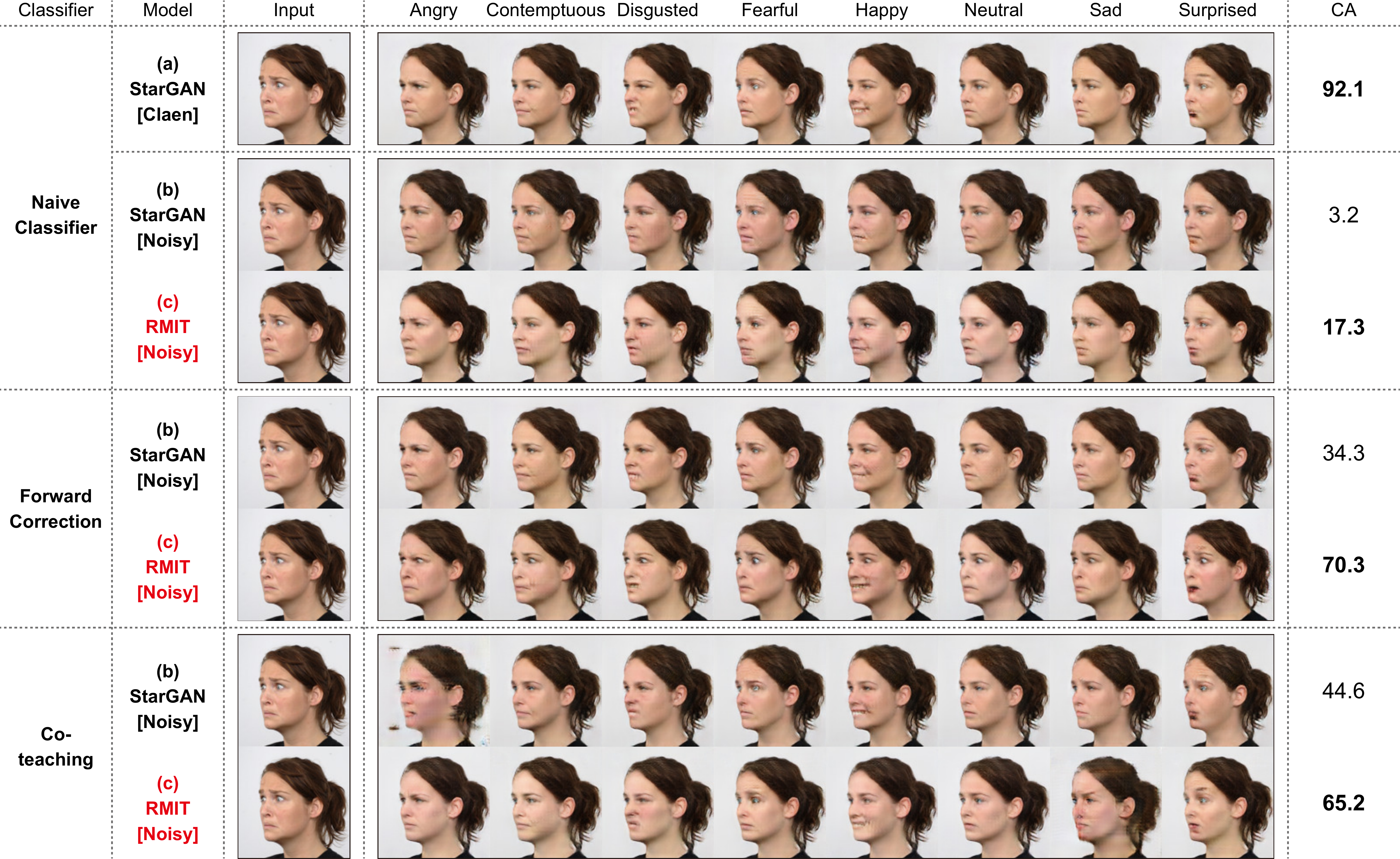}
  \caption{(Best zoomed in.) Examples of label-noise robust multi-domain image-to-image translation. This represents the extended results of Figure~\ref{fig:concept}. The view of the figure is the same as for Figure~\ref{fig:concept_ex1}.}
  \label{fig:concept_ex2}
\end{figure*}

\clearpage
\subsection{Extended results of Figure~\ref{fig:recon_error}}

\begin{figure*}[h]
  \centering
  \includegraphics[width=1\textwidth]{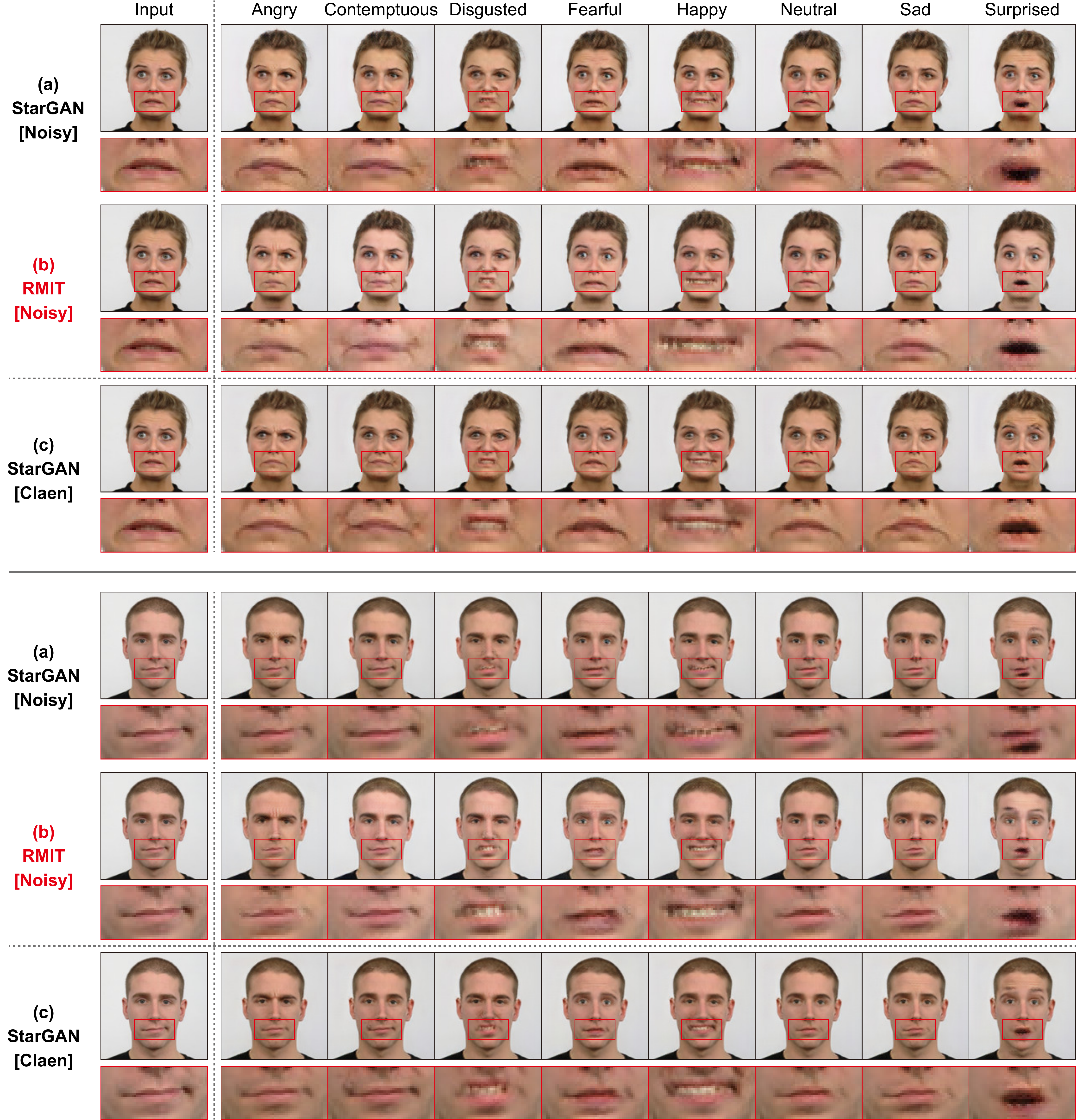}
  \caption{(Best zoomed in.) Generated images using models without a label-noise robust classifier (asymmetric noise with $\mu = 0.3$). This represents the extended results of Figure~\ref{fig:recon_error}. In (a), StarGAN in the noisy label setting partially preserves the input information that is unnecessary for the translated image. This causes mixture artifacts around the mouth (e.g., the first row and sixth column). This results in the degradation of the CA ($60.2\%$). In (b), RMIT mitigates this problem and the generated images are close to the images generated by StarGAN in the clean-label setting in (c). Owing to this improvement, RMIT achieves the higher CA ($78.9\%$).}
  \label{fig:recon_error_ex}
\end{figure*}

\clearpage
\subsection{Extended results of Figure~\ref{fig:celeba}}
\label{subsec:gen_celeba_ex}

\begin{figure*}[h]
  \centering
  \includegraphics[height=0.81\textheight]{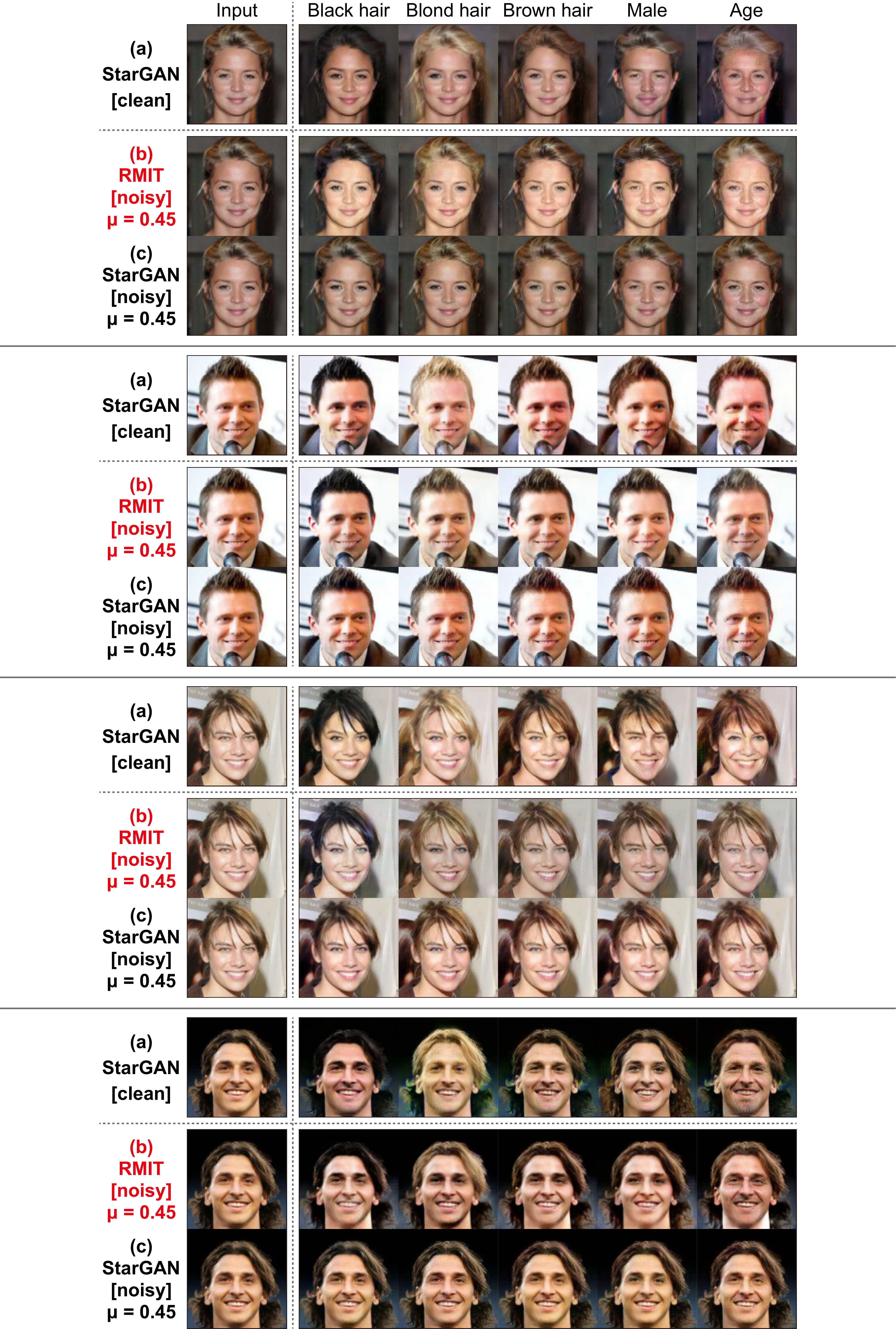}
  \caption{(Best zoomed in.) Generated images on CelebA. This represents the extended results of Figure~\ref{fig:celeba}. In (a), StarGAN in the clean-label setting learns a clean label conditional model. In (b), RMIT struggles to conduct meaningful conversion like StarGAN in the clean label setting (a). In contrast, in (c), StarGAN in the noisy label setting is adjacent to a nonconversion model. It is noteworthy that in each block (e.g., from the first to the third row), the images in the upper rows achieve a better CA but worse FID. Namely, in this dataset, StarGAN in the clean label setting achieves the best CA but is defeated by StarGAN in the noisy label setting in terms of the FID. As discussed in Figure~\ref{fig:ca_vs_fid}, this is because a nonconversion model can also achieve a high performance in terms of the FID even though the CA is worse. This finding indicates that balancing between the FID and CA is important for achieving good label-noise robust multi-domain image-to-image translation.}
  \label{fig:celeba_ex}
\end{figure*}

\clearpage
\subsection{Extended results of Figure~\ref{fig:fer}}
\label{subsec:gen_fer_ex}

\begin{figure*}[h]
  \centering
  \includegraphics[width=\textwidth]{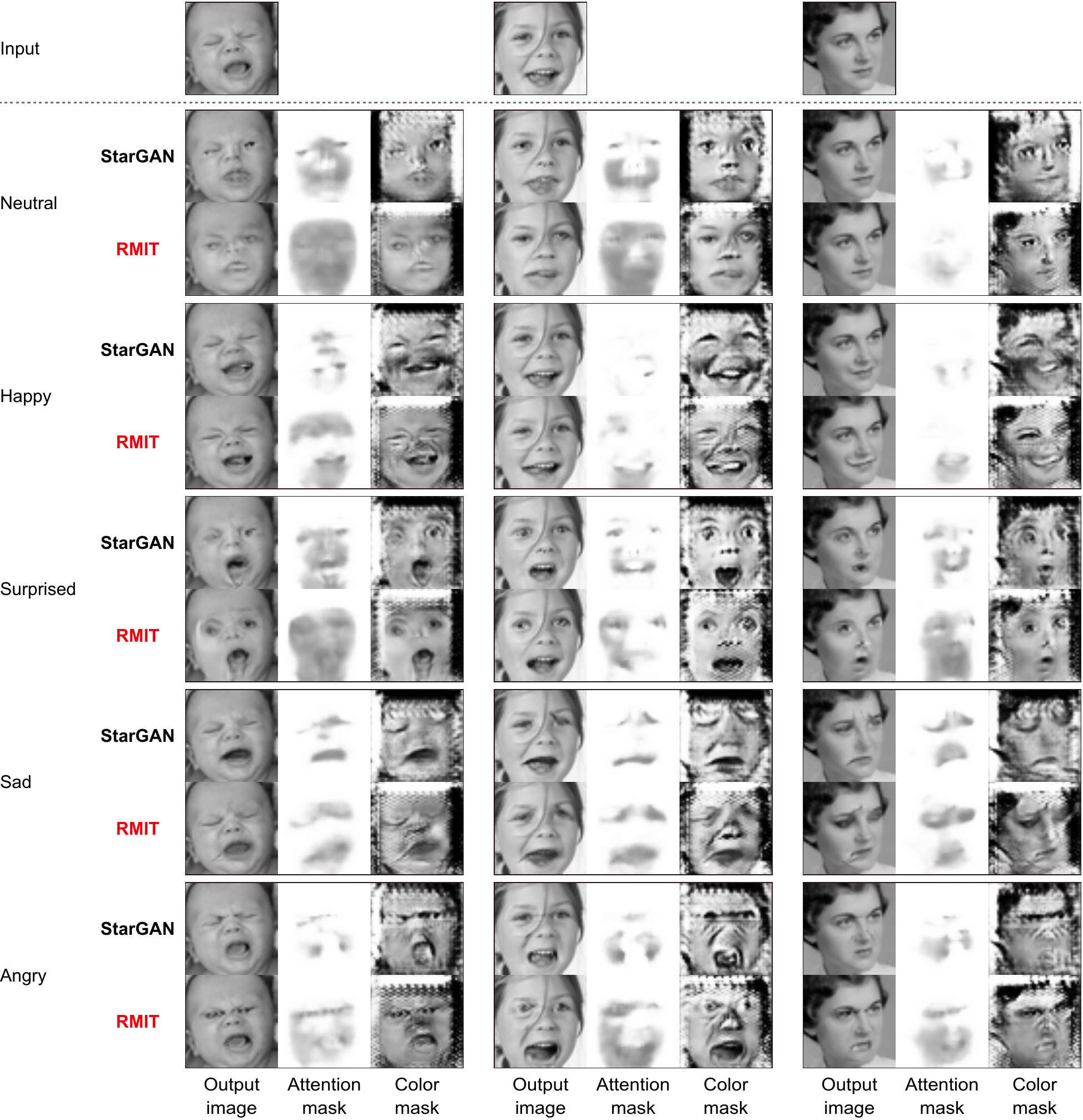}
  \caption{(Best zoomed in.) Generated images on FER. This represents the extended results of Figure~\ref{fig:fer}. The first row shows the input images ${\bm x}$. The remaining even rows include the images generated by StarGAN, and the remaining odd rows contain the images generated by RMIT. The first, fourth, and seventh columns show the output images ${\bm x}' = {\bm m} \cdot {\bm x}_{color} + (1 - {\bm m}) {\bm x}$, where ${\bm m}$ is the attention mask and ${\bm x}_{color}$ is the color mask. We present the attention masks ${\bm m}$ in the second, fifth, and eighth columns, and the color masks ${\bm x}_{color}$ in the third, sixth, and ninth columns. RMIT tends to learn larger attention masks (i.e., prefers larger variations) and generates more classifiable images than StarGAN. Indeed, RMIT achieves a CA of $70.0\%$, whereas StarGAN achieves a CA of $65.5\%$.}
  \label{fig:fer_ex}
\end{figure*}

\end{document}